\def\year{2018}\relax
\documentclass[letterpaper]{article} 
\usepackage{aaai18}  
\usepackage{times}  
\usepackage{helvet}  
\usepackage{courier}  
\usepackage{url}  
\usepackage{graphicx}  
\usepackage{bbm}
\usepackage{amsmath}
\usepackage{amsthm}
\usepackage{amssymb}
\usepackage{algorithm}
\usepackage{algorithmic}
\usepackage{caption}
\usepackage{subcaption}

\usepackage{multirow}
\usepackage[flushleft]{threeparttable}

\newtheorem{theorem}{Theorem}
\newtheorem{corollary}{Corollary}
\newtheorem{remark}{Remark}

\newtheorem{proposition}{Proposition}
\newtheorem{lemma}{Lemma}

\newtheorem{assumption}{Assumption}

\usepackage{color}
\newcommand{\blue}{}
\newcommand{\red}{}
\newcommand{\magenta}{}
\newcommand{\green}{}

\begin{document}
%
\title{A Change-Detection based Framework for\\ Piecewise-stationary Multi-Armed Bandit Problem}
\author{Fang Liu \and Joohyun Lee \and Ness Shroff\\
The Ohio State University\\
Columbus, Ohio 43210\\
\{liu.3977, lee.7119, shroff.11\}@osu.edu\\
}
\maketitle
\begin{abstract}
The multi-armed bandit problem has been extensively studied under the stationary assumption. However in reality, this assumption often does not hold because the distributions of rewards themselves may change over time. In this paper, we propose a change-detection (CD) based framework for multi-armed bandit problems under the piecewise-stationary setting, and study a class of change-detection based UCB (Upper Confidence Bound) policies, CD-UCB, that actively detects change points and restarts the UCB indices. We then develop CUSUM-UCB and PHT-UCB, that belong to the CD-UCB class and use cumulative sum (CUSUM) and Page-Hinkley Test (PHT) to detect changes. We show that CUSUM-UCB obtains the best known regret upper bound under mild assumptions. We also demonstrate the regret reduction of the CD-UCB policies over arbitrary Bernoulli rewards and Yahoo! datasets of webpage click-through rates.
\end{abstract}
\section{Introduction}
The multi-armed bandit problem, introduced by~\citeauthor{thompson1933likelihood}~\shortcite{thompson1933likelihood}, models sequential allocation in the presence of uncertainty and partial feedback on rewards. It has been extensively studied and has turned out to be fundamental to many problems in artificial intelligence, such as reinforcement learning~\cite{sutton1998reinforcement}, online recommendation systems~\cite{li2016collaborative} and computational advertisement~\cite{buccapatnam2017reward}. In the classical multi-armed bandit problem~\cite{lai1985asymptotically}, a decision maker needs to choose one of $K$ independent arms and obtains the associated reward in a sequence of time slots (rounds). Each arm is characterized by an unknown reward distribution and the rewards are independent and identically distributed (i.i.d.). 

The goal of a bandit algorithm, implemented by the decision maker, is to minimize the \emph{regret} over $T$ time slots, which is defined as the expectation of the difference between the total rewards collected by playing the arm with the highest expected reward and the total rewards obtained by the algorithm. To achieve this goal, the decision maker is faced with an {\em exploration versus exploitation} dilemma, which is the trade-off between exploring the environment to find the most profitable arms and exploiting the current empirically best arm as often as possible. A problem-dependent regret lower bound, $\Omega(\log T)$, of any algorithm for the classical bandit problem has been shown in~\citeauthor{lai1985asymptotically}~\shortcite{lai1985asymptotically}. Several algorithms have been proposed and proven to achieve $O(\log T)$ regret, such as Thompson Sampling~\cite{agrawal2012analysis}, $\epsilon_n$-greedy and Upper Confidence Bound (UCB)~\cite{auer2002finite}. Variants of these bandit policies can be found in~\citeauthor{bubeck2012regret}~\shortcite{bubeck2012regret}.

Although the stationary (classical) multi-armed bandit problem is well-studied, it is unclear whether it can achieve $O(\log T)$ regret in a non-stationary environment, where the distributions of rewards change over time. This setting often occurs in practical problems. For example, consider the dynamic spectrum access problem \cite{alaya2008dynamic} in communication systems. Here, the decision maker wants to exploit the empty channel, thus improving the spectrum usage. The availability of a channel is dependent on the number of users in the coverage area. 
The number of users, however can change dramatically with time of day and, therefore, is itself a non-stationary stochastic process.
Hence, the availability of the channel also follows a distribution that is not only unknown, but varies over time. 
To address the changing environment challenge, a non-stationary multi-armed bandit problem
has been proposed in the literature. There are two main approaches to deal with the non-stationary environment: \emph{passively adaptive policies}~\cite{garivier2008upper,besbes2014stochastic,wei2016tracking} and {\em actively adaptive policies}~\cite{hartland2007change,mellor2013thompson,allesiardo2015exp3}.

First, {\em passively adaptive policies} are unaware of when changes happen but update their decisions based on the most recent observations in order to keep track of the current best arm. Discounted UCB (D-UCB), introduced by~\citeauthor{kocsis2006discounted}~\shortcite{kocsis2006discounted}, where geometric moving average over the samples is applied to the UCB index of each arm, has been shown to achieve the regret upper-bounded by $O(\sqrt{\gamma_T T}\log T)$, where $\gamma_T$ is the number of change points up to time $T$~\cite{garivier2008upper}. Based on the analysis of D-UCB, they also proposed and analyzed Sliding-Window UCB (SW-UCB), where the algorithm updates the UCB index based on the observations within a moving window of a fixed length. The regret of SW-UCB is at most $O(\sqrt{\gamma_T T\log T})$. Exp3.S~\cite{doi:10.1137/S0097539701398375} also achieves the same regret bound, where a uniform exploration is mixed with the standard Exp3~\cite{cesa2006prediction} algorithm.
Similarly,~\citeauthor{besbes2014stochastic}~\shortcite{besbes2014stochastic} proposed a Rexp3 algorithm, which restarts the Exp3 algorithm periodically. It is shown that the regret is upper-bounded by $O(V_T^{1/3}T^{2/3})$, where $V_T$ denotes the total reward variation budget up to time $T$.\footnote{\blue $V_T$ satisfies $\sum_{t=1}^{T-1} \sup_{i \in \mathcal{K}} |\mu_t(i) - \mu_{t+1}(i)| \leq V_T$ for the expected reward of arm $i$ at time $t$, $\mu_t(i)$.} The increased regret of Rexp3 comes from the adversarial nature of the algorithm, which assumes that the environment changes every time slot
in the worst case.

Second, {\em actively adaptive policies} adopt a change detection algorithm to monitor the varying environment and restart the bandit algorithms when there is an alarm. Adapt-EvE, proposed by~\citeauthor{hartland2007change}~\shortcite{hartland2007change}, employs a Page-Hinkley Test (PHT)~\cite{hinkley1971inference} to detect change points and restart the UCB policy. PHT has also been used to adapt the window length of SW-UCL~\cite{srivastava2014surveillance}, which is an extension of SW-UCB in the multi-armed bandit with Gaussian rewards. However, the regret upper bounds of Adapt-EvE and adaptive SW-UCL are still open problems. These works are closely related to our work, as one can regard them as instances of our change-detection based framework. We highlight that one of our contributions is to provide an analytical result for such a framework. \citeauthor{mellor2013thompson}~\shortcite{mellor2013thompson} took a Bayesian view of the non-stationary bandit problem, where a stochastic model of the dynamic environment is assumed and a Bayesian online change detection algorithm is applied. Similar to the work by~\citeauthor{hartland2007change}~\shortcite{hartland2007change}, the theoretical analysis of the Change-point Thompson Sampling (CTS) is still open. Exp3.R~\cite{allesiardo2015exp3} combines Exp3 and a drift detector, and achieves the regret $O(\gamma_T \sqrt{T\log T})$, which is not efficient when the change rate $\gamma_T$ is high.

In sum, for various passively adaptive policies theoretical guarantees have been obtained, as they are considered more tractable to analyze. However, it has been demonstrated via extensive numerical studies that actively adaptive policies outperform passively adaptive policies~\cite{mellor2013thompson}. The intuition behind this is that actively adaptive policies can utilize the balance between exploration and exploitation by bandit algorithms, once a change point is detected and the environment stays stationary for a while, which is often true in real world applications. This observation motivates us to construct a change-detection based framework, where a class of actively adaptive policies can be developed with both good theoretical bounds and good empirical performance. Our main contributions are as follows.
\vspace{-3pt}
\begin{enumerate}
 \item We propose a change-detection based framework for a piecewise-stationary bandit problem, which consists of a change detection algorithm and a bandit algorithm. We develop a class of policies, CD-UCB, that uses UCB as a bandit algorithm. We then design two instances of this class, CUSUM-UCB and PHT-UCB, that exploit CUSUM (cumulative sum) and PHT as their change detection algorithms, respectively.
 \vspace{-5pt}
 \item 
We provide a regret upper bound for the CD-UCB class, for given change detection performance. For CUSUM, we obtain an upper bound on the mean detection delay and a lower bound on the mean time between false alarms, and show that the regret of CUSUM-UCB is at most $O(\sqrt{T\gamma_T\log{\frac{T}{\gamma_T}}})$. To the best of our knowledge, this is the first regret bound for actively adaptive UCB policies in the bandit feedback setting.
 \vspace{-5pt}
 \item 
The performance of the proposed and existing policies are validated by both synthetic and real world datasets, and we show that our proposed algorithms are superior to other existing policies in terms of regret.
 \end{enumerate}
We present the problem setting in Section~\ref{sec:problemform} and introduce our framework in Section~\ref{sec:framework}. We propose our algorithms in Section~\ref{sec:algorithm}. We then present performance guarantees in Section~\ref{sec:analysis}. In Section~\ref{sec:simul}, we compare our algorithms with other existing algorithms via simulation. Finally, we conclude the paper. 
\section{Problem Formulation}\label{sec:problemform}
\subsection{Basic Setting}
Let $\mathcal{K}=\{1,\ldots,K\}$ be a set of arms. Let $\{1,2,\ldots,T\}$ denote the decision slots faced by a decision maker and $T$ is the time horizon. At each time slot $t$, the decision maker chooses an arm $I_t\in\mathcal{K}$ and obtains a reward $X_t(I_t)\in[0,1]$. Note that the results can be generalized to any bounded interval. The rewards $\{X_t(i)\}_{t\geq1}$ for arm $i$ are modeled by a sequence of independent random variables from potentially different distributions, which are unknown to the decision maker. 
Let $\mu_t(i)$ denote the expectation of reward $X_t(i)$ at time slot $t$, i.e., $\mu_t(i)=\mathbb{E}[X_t(i)]$. Let $i^*_t$ be the arm with highest expected reward at time slot $t$, denoted by $\mu_t(*)\triangleq\mu_t(i^*_t)=\max_{i\in\mathcal{K}}\mu_t(i)$.
Let $\Delta_{\mu_T(i)}\triangleq\min\{\mu_t(*)-\mu_t(i):t\leq T, i\neq i^*_t\}$, be the minimum difference over all time slots between the expected rewards of the best arm $i^*_t$ and the arm $i$ while the arm $i$ is not the best arm. 

A policy $\pi$ is an algorithm that chooses the next arm to play based on the sequence of past plays and obtained rewards. The performance of a policy $\pi$ is measured in terms of the \emph{regret}. The regret of $\pi$ after $T$ plays is defined as the expected total loss of playing suboptimal arms. Let $R_\pi(T)$ denote the regret of policy $\pi$ after $T$ plays and let $\tilde{N}_T(i)$ be the number of times arm $i$ has been played when it is not the best arm by $\pi$ during the first $T$ plays.
\begin{align}\label{def:nti}
R_\pi(T)&=\mathbb{E}\left[\sum_{t=1}^T\left(X_t(i^*_t)-X_t(I_t)\right)\right],\\
\tilde{N}_T(i)&=\sum_{t=1}^T\mathbbm{1}_{\left\{I_t=i,~\mu_t(i)\neq \mu_t(*)\right\}}.
\end{align}
Note that the regret $R_\pi(T)$ of policy $\pi$ is upper-bounded by $\sum_{i=1}^K\mathbb{E}[\tilde{N}_T(i)]$ since the rewards are bounded in (\ref{def:nti}). 
In Section~\ref{sec:analysis}, we provide an upper bound on $\mathbb{E}[\tilde{N}_T(i)]$ to obtain a regret upper bound.

\subsection{Piecewise-stationary Environment}
We consider the notion of a \emph{piecewise-}stationary environment in~\citeauthor{yu2009piecewise}~\shortcite{yu2009piecewise}, where the distributions of rewards remain constant for a certain period and abruptly change at some unknown time slots, called \emph{breakpoints}. Let $\gamma_T$ be the number of breakpoints up to time $T$, $\gamma_T=\sum_{t=1}^{T-1}\mathbbm{1}_{\{\exists i\in\mathcal{K}: \mu_t(i)\neq \mu_{t+1}(i)\}}.$
In addition, we make three mild assumptions for tractability.
\begin{assumption}\label{ass:piecewise}
\emph{(Piecewise Stationarity)} The shortest interval between two consecutive breakpoints is greater than $K M$, for some integer $M$.
\end{assumption}
Assumption \ref{ass:piecewise} ensures that the shortest interval between two successive breakpoints is greater than $K M$, so that we have enough samples to estimate the mean of each arm before the change happens. Note that this assumption is equivalent to the notions of an \emph{abruptly changing environment} used in~\citeauthor{garivier2008upper}~\shortcite{garivier2008upper} and a \emph{switching} environment in~\citeauthor{mellor2013thompson}~\shortcite{mellor2013thompson}. However, it is different from the adversarial environment assumption, where the environment changes all the time.
We make a similar assumption as Assumption 4.2 in~\citeauthor{yu2009piecewise}~\shortcite{yu2009piecewise} about the detectability in this paper.
\begin{assumption}\label{ass:detectability}
\emph{(Detectability)} There exists a known parameter $\epsilon>0$, such that $\forall i\in\mathcal{K}$ and $\forall t\leq T-1$, if $\mu_t(i)\neq\mu_{t+1}(i)$, then $|\mu_t(i)-\mu_{t+1}(i)|\geq 3\epsilon$.
\end{assumption}
Assumption \ref{ass:detectability} excludes infinitesimal mean shift, which is reasonable in practice when detecting abrupt changes bounded from below by a certain threshold.

\begin{assumption}\label{ass:lambda}
\emph{(Bernoulli Reward)} The distributions of all the arms are Bernoulli distributions.
\end{assumption}
Assumption \ref{ass:lambda} has also been used in the literature~\cite{besbes2014stochastic,mellor2013thompson,kaufmann2012thompson,agrawal2012analysis}. By Assumption \ref{ass:lambda}, the empirical average of $M$ Bernoulli random variables must be one of the grid points $\{0,1/M,\ldots,1\}$. Let $\lambda_T(i)=\min\{(\mu_t(i)-\epsilon)-{\lfloor (\mu_t(i)-\epsilon)M\rfloor}/{M}, {\lceil (\mu_t(i)+\epsilon)M\rceil}/{M}-(\mu_t(i)+\epsilon): t \leq T\}\setminus\{0\}$ be the minimal non-trival gap between expectation and closest grid point of arm $i$.\footnote{Note that $\lfloor\cdot\rfloor$ denotes the floor function and $\lceil\cdot\rceil$ denotes the ceiling function.} We define the minimal gap of all arms as $\lambda=\min_{i\in\mathcal{K}}\lambda_T(i)$.
\section{Change-Detection based Framework}\label{sec:framework}
\begin{figure}
  \centering
    \includegraphics[width=0.35\textwidth]{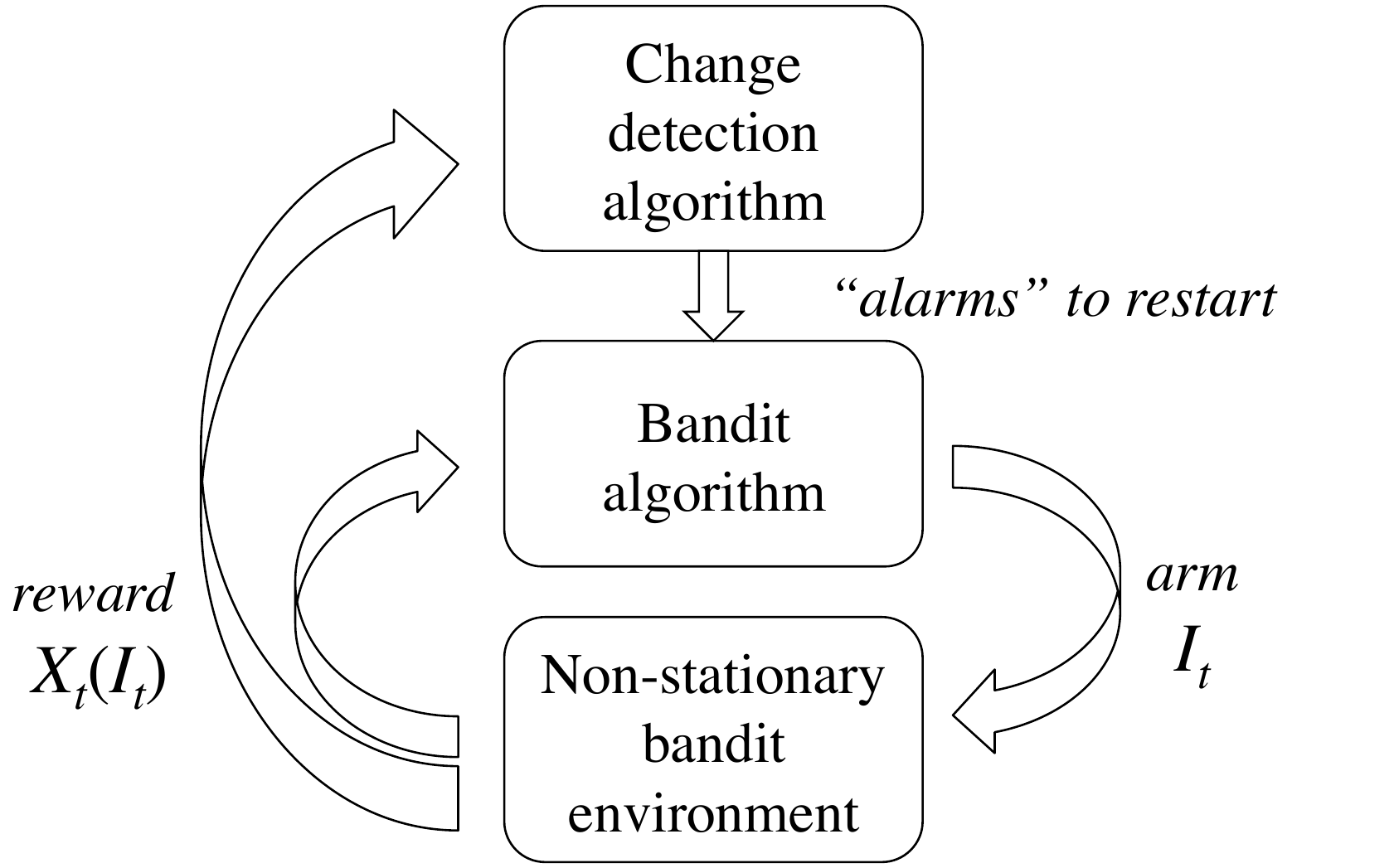} 
     \caption{Change-detection based framework for non-stationary bandit problems}
     \label{fig:framework}
\end{figure}

Our change-detection based framework consists of two components: a change detection algorithm and a bandit algorithm, as shown in Figure~\ref{fig:framework}. At each time $t$, the bandit algorithm outputs a decision $I_t\in\mathcal{K}$ based on its past observations of the bandit environment. The environment generates the corresponding reward of arm $I_t$, which is observed by both the bandit algorithm and the change detection algorithm. The change detection algorithm monitors the distribution of each arm, and sends out a positive signal to restart the bandit algorithm once a breakpoint is detected. One can find that our framework is a generalization of the existing actively adaptive policies. 


Since the bandit algorithms are well-studied in the bandit setting, what remains is to find a change detection algorithm, which works in the bandit environment. Change point detection problems have been well studied, see, e.g., the book~\cite{basseville1993detection}. However, the change detection algorithms are applied in a context that is quite different from the bandit setting. There are two key challenges in adapting the existing change detection algorithms in the bandit setting.

(1) \emph{Unknown priors:} In the context of the change detection
problem, one usually assumes that the prior distributions before and
after a change point are known. However, such information is unknown
to the decision maker in the bandit setting. Even though there are
some simple methods, such as estimating the priors and then applying
the change detection algorithm {\blue like PHT}, there are no analytical results in the
literature.

(2) \emph{Insufficient samples:} Due to the bandit feedback setting,
the decision maker can only observe one arm at each time. However,
there are $K$ change detection algorithms running in parallel since
each arm is associated with a change detection procedure to monitor
the possible mean shift. So the change detection algorithms in most
arms are hungry for samples at each time. If the decision maker does
not feed these change detection algorithms intentionally, the change
detection algorithm may miss detection opportunities because they do
not have enough recent samples.


\section{Application of the Framework}\label{sec:algorithm}
\begin{figure*}
  \centering
\begin{minipage}{.49\textwidth}
\begin{algorithm}[H]
\caption{CD-UCB}
\label{alg:CD-UCB}
\begin{algorithmic}
\REQUIRE $T$, $\alpha$ and an algorithm CD$(\cdot,\cdot)$
\STATE Initialize $\tau_i=1, \forall i$.
\FOR{$t$ {\bfseries from} $1$ {\bfseries to} $T$}
\STATE{Update according to equations (\ref{eqn:update}-\ref{eqn:It}).}
\STATE{{\blue Play arm $I_t$ and observe $X_t(I_t)$.}}
\IF{CD$(I_t,X_t(I_t))==1$}
\STATE{$\tau_{I_t}=t+1$; reset CD$(I_t,\cdot)$.}
\ENDIF
\ENDFOR
\end{algorithmic}
\end{algorithm}
\end{minipage} \hfill
\begin{minipage}{.49\textwidth}
  \begin{algorithm}[H]
 \caption{Two-sided CUSUM}
\label{alg:CUSUM}
\begin{algorithmic}
\REQUIRE parameters $\epsilon$, $M$, $h$ and $\{y_k\}_{k\geq 1}$
\STATE Initialize $g_0^+=0$ and $g_0^-=0$.
\FOR{each $k$}
\STATE Calculate $s_k^-$ and $s_k^+$ according to (\ref{twoCUSUM}).
\STATE Update $g_k^+$ and $g_k^-$ according to (\ref{eqn:g+-}). 
\IF{$g_k^+\geq h$ or $g_k^-\geq h$}
\STATE Return 1
\ENDIF
\ENDFOR
\end{algorithmic}
\end{algorithm}
\end{minipage} 
\end{figure*}

In this section, we introduce our Change-Detection based UCB (CD-UCB) policy, which addresses the issue of insufficient samples. Then we develop a tailored CUSUM algorithm for the bandit setting to overcome the issue of unknown priors. Finally, we combine our CUSUM algorithm with the UCB algorithm as CUSUM-UCB policy, which is a specific instance of our change-detection based framework. Performance analysis is provided in Section~\ref{sec:analysis}.

\subsection{CD-UCB policy}
Suppose we have a change detection algorithm, CD$(\cdot,\cdot)$, which takes arm index $i$ and observation $X_t(i)$ as input at time $t$, and it returns $1$ if there is an alarm for a breakpoint. Given such a change detection algorithm, we can employ it to control the UCB algorithm, which is our CD-UCB policy as shown in Algorithm~\ref{alg:CD-UCB}. We clarify some useful notations as follows. 
{\magenta Let $\tau_i = \tau_i(t)$ be the last time that the CD$(i,\cdot)$ alarms and restarts for arm $i$ before time $t$.}
Then the number of valid observations (after the latest detection alarm) for arm $i$ up to time $t$ is denoted as $N_t(i)$. Let $n_t$ be the total number of valid observations for the decision maker. For each arm $i$, let $\bar{X}_t(i)$ be the sample average and $C_t(i)$ be the confidence padding term. In particular,
\begin{align}\label{eqn:update}
N_t(i)&=\sum_{s=\tau_i}^t\mathbbm{1}_{\{I_s=i\}},~~~n_t=\sum_{i=1}^K N_t(i),\\
\bar{X}_t(i) &\!=\! \sum_{s=\tau_i}^t\frac{X_s(i)}{N_t(i)}\mathbbm{1}_{\{I_s=i\}},~~~C_t(i) \!=\!  \sqrt{\frac{\xi\log{n_t}}{N_t(i)}},
\end{align}
where $\xi$ is some positive real number.
Thus, the UCB index for each arm $i$ is $\bar{X}_t(i)+C_t(i)$. Parameter $\alpha$ is a tuning parameter we introduce in the CD-UCB policy. At each time $t$, the policy plays the arm 
\begin{eqnarray}\label{eqn:It}
I_t=
\begin{cases}
\arg\max_{i\in\mathcal{K}}\left(\bar{X}_t(i)+C_t(i)\right),  &\text{w.p. } 1-\alpha\cr
i, & \forall i \in \mathcal{K}, \text{w.p. } \frac{\alpha}{K}
\end{cases}.
\end{eqnarray}


Parameter $\alpha$ 
controls the fraction of plays we exploit to feed the change detection algorithm. A large $\alpha$ may drive the algorithm to a linear regret performance while a small $\alpha$ can limit the detectability of change detection algorithm. We will discuss the choice of $\alpha$ in Sections~\ref{sec:analysis} and~\ref{sec:simul}.

\subsection{Tailored CUSUM algorithm}

A change detection algorithm observes a sequence of independent random variables, $y_1,y_2,\ldots$, in an online manner, and outputs an alarm once a change point is detected. In the context of the traditional change detection problem, one assumes that the parameters $\theta_0$ and $\theta_1$ are known for the density function $p(\cdot|\theta)$. In addition, $y_k$ is sampled from distribution under $\theta_0$ ($\theta_1$) before (after) the breakpoint. Let $u_0$ ($u_1$) be the mean of $y_k$ before (after) the change point. The CUSUM algorithm, originally proposed by~\cite{page1954continuous}, has been proven to be optimal in detecting abrupt changes in the sense of worst mean detection delay~\cite{lorden1971procedures}. 
{\blue The basic idea of the CUSUM algorithm is to take a function of the observed sample (e.g., the logarithm of likelihood ratio $\log\frac{p(y_k|\theta_1)}{p(y_k|\theta_0)}$) as the step of a random walk.}
{\blue This random walk is designed to have a positive mean drift after a change point and have a negative mean drift without a change.}
Hence, {\blue CUSUM signals a change if} this random walk crosses some positive threshold $h$.

We propose a tailored CUSUM algorithm that works in the bandit setting.
To be specific, we use the first $M$ samples to calculate the average, $\hat{u}_0\triangleq({\sum_{k=1}^My_k})/{M}.$
Then we construct two random walks, which have negative mean drifts before the change point and have positive mean drifts after the change. In particular, we design a two-sided CUSUM algorithm, described in Algorithm~\ref{alg:CUSUM}, with an upper (lower) random walk monitoring the possible positive (negative) mean shift. Let $s_k^+$ ($s_k^-$) be the step of the upper (lower) random walk. Then $s_k^+$ and $s_k^-$ are defined as
\begin{equation}\label{twoCUSUM}
(s_{k}^+,s_{k}^-)=(y_k-\hat{u}_0-\epsilon,\hat{u}_0-y_k-\epsilon)\mathbbm{1}_{\{k>M\}}.
\end{equation}
Let $g_k^+$ ($g_k^-$) track the positive drift of upper (lower) random walk. In particular,
\begin{align}
\label{eqn:g+-}
g_k^+=\max(0,g_{k-1}^++s_k^+), \,\, g_k^-=\max(0,g_{k-1}^-+s_k^-).
\end{align}
The change point is detected when either of them crosses the threshold
$h$. The parameter $h$ is important in the detection delay and
false alarm trade-off. We discuss the choice of $h$ in Section~\ref{sec:analysis}.

\subsection{CUSUM-UCB policy}

Now we are ready to introduce our CUSUM-UCB policy, which is a CD-UCB policy with CUSUM as a change detection algorithm. In particular, it takes $K$ parallel CUSUM algorithms as CD$(\cdot,\cdot)$ in CD-UCB. Formal description of CUSUM-UCB can be found in Algorithm~\ref{alg:CUSUM-UCB}, provided in Section \ref{app:cusumucb} in the supplementary material. 

\subsection{PHT-UCB policy}
We introduce another instance of our CD-UCB with the PHT algorithm~\cite{hinkley1971inference} running as the change detection algorithm, named PHT-UCB. PHT can be viewed as a variant of Algorithm~\ref{alg:CUSUM} by replacing (\ref{twoCUSUM}) with $(s_{k}^+,s_{k}^-)=(y_k-\hat{y}_k-\epsilon,\hat{y}_k-y_k-\epsilon),$
where $\hat{y}_k=\frac{1}{k}\sum_{s=1}^ky_s$. 
\section{Performance Analysis}\label{sec:analysis}

{In this section, we analyze the performance in each part of the proposed
algorithm: (a) our bandit algorithm (i.e., CD-UCB), and (b) our change
detection algorithm (i.e., two-sided CUSUM).  First, we present the
regret upper bound result of CD-UCB for a given change detection
guarantee.
This is of independent interest in understanding the
challenges of the non-stationary environment.  Second, we provide 
performance guarantees of our modified CUSUM algorithm in terms of the
mean detection delay, $\mathbb{E}[D]$, and the expected number of false alarms up to time $T$, $\mathbb{E}[F]$. Then, we combine
these two results to provide the regret upper bound of our CUSUM-UCB.}
{\blue The proofs are presented in our supplementary material.}


\begin{theorem}\label{thm:CD-UCB}
\emph{(CD-UCB)} Let $\xi=1$. Under Assumption~\ref{ass:piecewise}, for any $\alpha\in[0,1)$ and any arm $i\in\{1,\ldots,K\}$, the CD-UCB policy achieves,
\begin{eqnarray}\label{thm1}
\mathbb{E}[\tilde{N}_T(i)]\leq  
&\left(\gamma_T+\mathbb{E}[F]\right)\cdot\left(\frac{4\log T}{(\Delta_{\mu_T(i)})^2}+\pi^2/3\right) \\\nonumber
&+\frac{\pi^2}{3}
+\gamma_T\cdot \mathbb{E}[D]+\frac{\alpha T}{K}.
\end{eqnarray}
\end{theorem}

Recall that the regret of the CD-UCB policy is upper-bounded by
$\sum_{i=1}^K\mathbb{E}[\tilde{N}_T(i)]$. 
{\blue Therefore, given the parameter values (e.g., $\alpha$) and
 the performance of a change detection algorithm (i.e.,
 $\mathbb{E}[F]$ and $\mathbb{E}[D]$),
we can obtain the regret upper bound of that change detection based
bandit algorithm.}
{\blue By letting $\alpha=0$, we obtain the following result.} 

\begin{corollary}\label{thm:0CD-UCB}
\emph{(CD-UCB$|\alpha=0$)} If $\alpha=0$ and $\xi=1$, then the regret of CD-UCB is
\begin{equation}
R_{\pi^{\text{CD-UCB}}}(T)=O((\gamma_T+\mathbb{E}[F])\cdot\log T+\gamma_T\cdot\mathbb{E}[D])).
\end{equation}
\end{corollary}
\begin{remark}
If one can find an oracle algorithm that detects the change point with
the properties that $\mathbb{E}[F]\leq O(\gamma_T)$ and
$\mathbb{E}[D]\leq O(\log T)$, then one can achieve $O(\gamma_T \log
T)$ regret, which recovers the regret result
in~\citeauthor{yu2009piecewise}~\shortcite{yu2009piecewise}. We note that the WMD (Windowed Mean-shift Detection) change detection algorithm proposed by~\citeauthor{yu2009piecewise}~\shortcite{yu2009piecewise} achieves these properties when side observations are available.
\end{remark}
In the next proposition, we introduce the result of Algorithm
\ref{alg:CUSUM} about the conditional expected detection delay and the
conditional expected number of false alarms given $\hat{u}_0$. Note
that the expectations exclude the first $M$ slots for initial observations.
\begin{proposition}\label{prop:CUSUM}
\emph{(CUSUM$|\hat{u}_0$)}
Recall that $h$ is the tuning parameter in Algorithm~\ref{alg:CUSUM}.
Under Assumptions~\ref{ass:piecewise} and \ref{ass:detectability},
the conditional expected detection delay $\mathbb{E}\left[D\middle||\hat{u}_0-u_0|<\epsilon\right]$ and the conditional expected number of false alarms $\mathbb{E}\left[F\middle||\hat{u}_0-u_0|<\epsilon\right]$ satisfy
\begin{align}
\mathbb{E}\left[D\middle||\hat{u}_0-u_0|<\epsilon\right]\leq&\frac{h+1}{|u_1-\hat{u}_0|-\epsilon},\\
\mathbb{E}\left[F\middle||\hat{u}_0-u_0|<\epsilon\right]\leq&\frac{2T}{\exp(r(\theta_0)h)},
\end{align}
where $r(\theta_0)=\min(r^-(\theta_0),r^+(\theta_0))$, $r^-(\theta_0)$ is the non-zero root of $\log\mathbb{E}_{\theta_0}[e^{rs^-_{M+1}}]$ and $r^+(\theta_0)$ is the non-zero root of $\log\mathbb{E}_{\theta_0}[e^{rs^+_{M+1}}]$.
In the case of $|\hat{u}_0-u_0|>\epsilon$, the algorithm restarts in at most $\frac{h+1}{|\hat{u}_0-u_0|-\epsilon}$ time slots.
\end{proposition}

In the next theorem, we show the result for $\mathbb{E}[D]$ and
$\mathbb{E}[F]$ when CUSUM is used to detect the abrupt change. Note
again that the expectations exclude the first $M$ time slots.
\begin{theorem}\label{thm:CUSUM}
\emph{(CUSUM)} Under Assumptions~\ref{ass:piecewise}, \ref{ass:detectability} and \ref{ass:lambda}, the expected detection delay $\mathbb{E}[D]$ and the expected number of false alarms $\mathbb{E}[F]$ of the Algorithm \ref{alg:CUSUM} satisfy
\begin{align}
\mathbb{E}[D]\leq&{C_2(h+1)},\\
\mathbb{E}[F]\leq&\frac{2T}{(1-2\exp(-2\epsilon^2M))\exp(C_1h)},
\end{align}
where $C_2\triangleq\log(3)+2\exp(-2\epsilon^2M)/\lambda$, $C_1^-\triangleq\log\left(\frac{4\epsilon}{(1-\epsilon)^2}\binom{M}{\lfloor 2\epsilon M\rfloor}(2\epsilon)^M+1\right)$, $C_1^+\triangleq\log\left(\frac{4\epsilon}{(1+\epsilon)^2}\binom{M}{\lceil 2\epsilon M\rceil}(2\epsilon)^M+1\right)$ and $C_1\triangleq\min(C_1^-,C_1^+)$.
\end{theorem}

Summing the result of Theorems~\ref{thm:CD-UCB} and \ref{thm:CUSUM}, we obtain the regret upper bound of the CUSUM-UCB policy. To the best of our knowledge, this is the first regret bound for an actively adaptive UCB policy in the bandit feedback setting.
\begin{theorem}\label{thm:CUSUM-UCB}
\emph{(CUSUM-UCB)} Let $\xi=1$.  Under 
Assumptions~\ref{ass:piecewise}, \ref{ass:detectability} and
\ref{ass:lambda}, for any $\alpha\in(0,1)$ and any arm
$i\in\{1,\ldots,K\}$, the CUSUM-UCB policy achieves,
\begin{align}
\mathbb{E}[\tilde{N}_T(i)]\leq R_1\cdot
  R_2+\frac{\pi^2}{3}+\frac{\alpha T}{K},
\end{align}
\vspace{-0.2cm}
\begin{align*}
\text{for} \,\,\, R_1&=\gamma_T+\frac{2 T}{(1-2\exp(-2\epsilon^2M))\exp(C_1h)},\\
R_2&=\frac{4\log T}{(\Delta_{\mu_T(i)})^2}+\frac{\pi^2}{3}+M+\frac{C_2(h+1)K}{\alpha}.
\end{align*}
\end{theorem}

\begin{table*}[t]
  \small
  \centering
  \caption{
  Comparison of regret bounds in various algorithms.
  }
  \label{table:regret}
  \tabcolsep 1pt
\begin{center}
    \begin{tabular}{|c|c|c|c|c|c|c|}
    \hline
    & \multicolumn{3}{c|}{\em Passively adaptive} &
                                                   \multicolumn{2}{c|}{\em
                                                   Actively adaptive}
      & \\
    \hline
    \multirow{2}{*}{Policy} & D-UCB & SW-UCB & Rexp3
    & Adapt-EvE  & CUSUM-UCB & {\em lower bound}\\
    &  {\scriptsize\cite{kocsis2006discounted}}& {\scriptsize\cite{garivier2008upper}}& {\scriptsize\cite{besbes2014stochastic}} & {\scriptsize\cite{hartland2007change}}  & & {\scriptsize \cite{garivier2008upper}}\\
    \hline
    Regret & $O(\sqrt{T\gamma_T}\log T)$ & $O(\sqrt{T\gamma_T\log T})$ & $O(V_T^{1/3}T^{2/3})$
    & Unknown & $O(\sqrt{T\gamma_T\log{\frac{T}{\gamma_T}}})$ & $\Omega(\sqrt{T})$\\
    \hline
  \end{tabular}
\end{center}
  \label{tbl:regret_comparison}
\vspace{-0.0cm}
\end{table*}

\begin{corollary}\label{cor:regret}
Under the Assumptions~\ref{ass:piecewise}, \ref{ass:detectability} and \ref{ass:lambda},
if horizon $T$ and the number of breakpoints $\gamma_T$ are known in advance, then we can choose $h=\frac{1}{C_1}\log \frac{T}{\gamma_T}$ and $\alpha=K\sqrt{\frac{C_2\gamma_T}{C_1T}\log\frac{T}{\gamma_T}}$ so that
\begin{equation}
R_{\pi^{\text{CUSUM-UCB}}}(T)=O\left(\frac{\gamma_T\log T}{(\Delta_{\mu_T(i)})^2}+\sqrt{T\gamma_T\log\frac{T}{\gamma_T}}\right).
\end{equation}
\end{corollary}
{\green
\begin{remark}
The choices of parameters depend on the knowledge of $\gamma_T$. This is common in the non-stationary bandit literature. For example, the discounting factor of D-UCB and sliding window size of SW-UCB depend on the knowledge of $\gamma_T$. The batch size of Rexp3 depends on the knowledge of $V_T$, which denotes the total reward variation. It is practically viable when the reward change rate is regular such that one can accurately estimate $\gamma_T$ based on history.
\end{remark}
}
\begin{remark}
As shown in~\citeauthor{garivier2008upper}~\shortcite{garivier2008upper}, the lower bound of the problem
is $\Omega(\sqrt{T})$. Our policy approaches the optimal regret rate
in an order sense.
\end{remark}
\begin{remark}\label{rmk:delta}
For the SW-UCB policy, the regret analysis result is $R_{\pi^{\text{SW-UCB}}}(T)=O\left(\frac{\sqrt{T\gamma_T\log T}}{(\Delta_{\mu_T(i)})^2}\right)$~\cite{garivier2008upper}. If $\Delta_{\mu_T(i)}$ is a constant with respect to $T$, then $\sqrt{T\gamma_T\log T}$ term dominates and our policy achieves the same regret rate as SW-UCB. If $\Delta_{\mu_T(i)}$ goes to $0$ as $T$ increases, then the regret of CUSUM-UCB grows much slower than SW-UCB.
\end{remark}

Table \ref{table:regret} summarizes the regret upper bounds
of the existing and proposed algorithms in the non-stationary setting
when $\Delta_{\mu_T(i)}$ is a constant in $T$. Our policy has a smaller regret term with respect to $\gamma_T$ compared to SW-UCB. 

\section{Simulation Results}\label{sec:simul}
We evaluate the
{\blue existing and proposed policies 
in three non-stationary environments:
two synthetic dataset (flipping and switching scenarios) and one
real-world dataset from Yahoo!~\cite{YahooWebScope}.
Yahoo! dataset collected user click traces for news articles.}
Our PHT-UCB is similar to Adapt-EvE, but they are different in that Adapt-EvE ignores the issue of insufficient samples and includes other {\blue heuristic} methods dealing with the detection points. 

{\green
In the simulation, the parameters $h$ and $\alpha$ are tuned around
$h=\log(T/\gamma_T)$ and
$\alpha=\sqrt{\frac{\gamma_T}{T}\log(T/\gamma_T)}$ based on the
flipping environment. We suggest the practitioners to take the same approach because the choices of $h$ and $\alpha$ in Corollary \ref{cor:regret} are minimizing the regret upper bound rather than the regret. We use the same parameters $h$ and $\alpha$ for
CUSUM-UCB and PHT-UCB to compare the performances of CUSUM and
PHT. Parameters are listed in Table \ref{table:simul} in Section \ref{app:simul} of the appendix. Note that $\epsilon$ and $M$ are obtained based on the prior knowledge of the datasets.
The baseline algorithms are tuned similarly with the knowledge of $\gamma_T$ and $T$.
We take the average regret over 1000 trials for the synthetic dataset.
}

\subsection{Synthetic Datasets}
\begin{figure}[t]
\centering
\begin{subfigure}[b]{0.45\textwidth}
    \includegraphics[width=\textwidth]{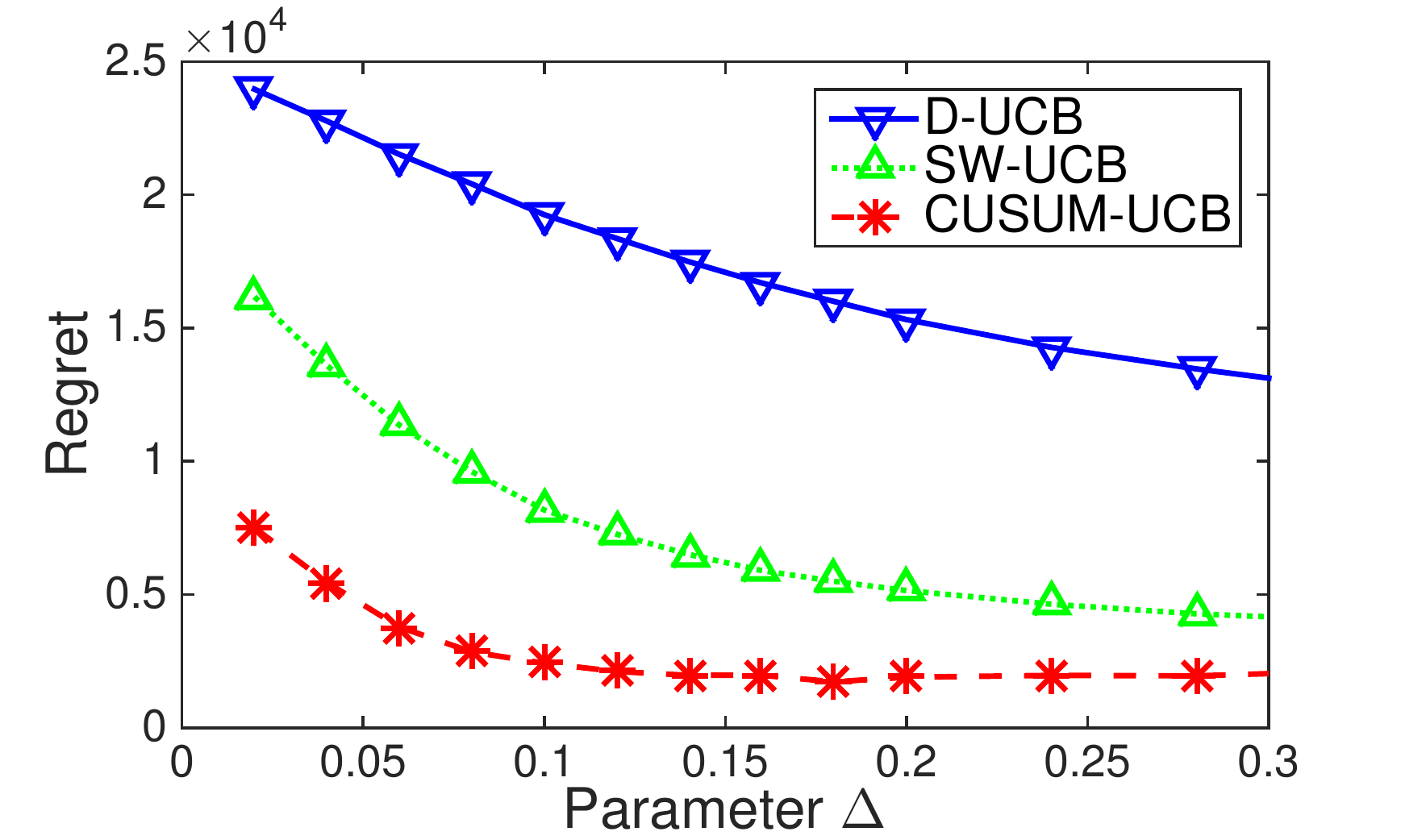}
     \caption{
 Under the flipping environment}\label{subfig:flip}
\end{subfigure}
\begin{subfigure}[b]{0.45\textwidth}
    \includegraphics[width=\textwidth]{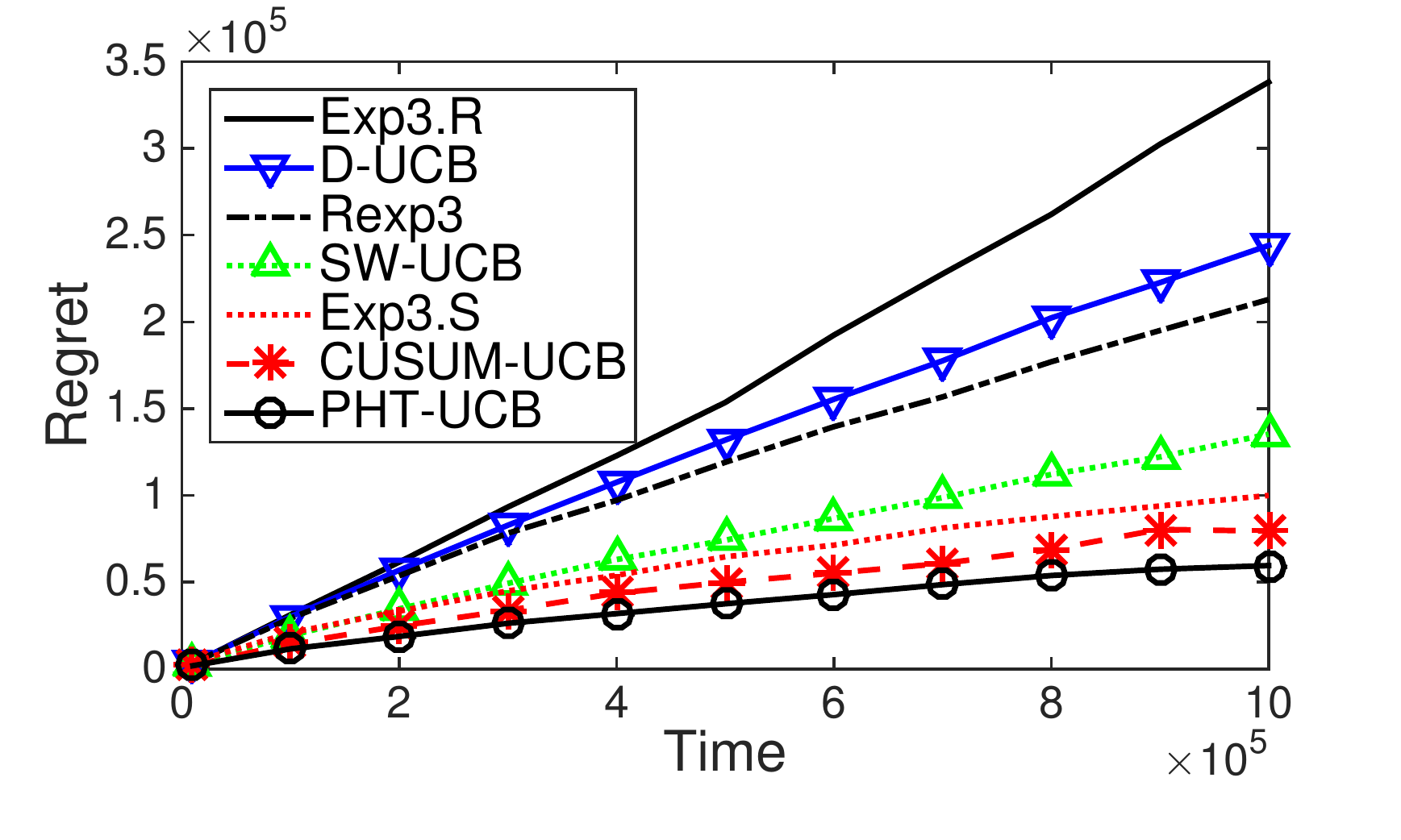}
     \caption{
Under the switching environment}\label{subfig:switch}
\end{subfigure}
\caption{Regret over synthetic datasets}
\label{fig:synthetic}
\end{figure}

{\bf Flipping Environment.}
We consider two arms (i.e., $K=2$) in the flipping environment, where arm $1$ is stationary and the expected reward of arm $2$ flips between two values. All arms are associated with Bernoulli distributions. In particular, $\mu_t(1)=0.5$ for any $t\leq T$ and 
\begin{eqnarray}\label{eqn:flip}
\mu_t(2)=
\begin{cases}
0.5-\Delta, &\frac{T}{3}\leq t\leq\frac{2T}{3}\cr
0.8, &\text{otherwise}
\end{cases}.
\end{eqnarray}
The two change points are at $\frac{T}{3}$ and $\frac{2T}{3}$. Note that $\Delta$ is equivalent to $\Delta_{\mu_T(2)}$.
We let $\Delta$ vary within the interval $[0.02,0.3]$, and compare the
regrets of D-UCB, SW-UCB and CUSUM-UCB to verify Remark~\ref{rmk:delta}. For this reason, results of other algorithms are omitted.
As shown in Figure~\ref{subfig:flip}, CUSUM-UCB outperforms D-UCB and SW-UCB. In addition, the gap between CUSUM-UCB and SW-UCB increases as $\Delta$ decreases.


{\bf Switching Environment.}
We consider the switching environment, introduced by~\citeauthor{mellor2013thompson}~\shortcite{mellor2013thompson}, which is defined by a hazard function, $\beta(t)$, such that,
\begin{eqnarray}\label{eqn:switch}
\mu_t(i)=
\begin{cases}
\mu_{t-1}(i), &\text{with probability } 1-\beta(t)\cr
\mu\sim U[0,1], &\text{with probability } \beta(t)
\end{cases}.
\end{eqnarray}
Note that $U[0,1]$ denotes the uniform distribution over the interval $[0,1]$ and $\mu_0(i)$ are independent samples from $U[0,1]$. In the experiments, we use the constant hazard function $\beta(t)=\gamma_T/T$. All the arms are associated with a Bernoulli distribution.

The regrets over the time horizon are shown in
Figure~\ref{subfig:switch}.  Although
Assumptions~\ref{ass:piecewise} and \ref{ass:detectability} are violated, CUSUM-UCB and PHT-UCB outperform the other policies. 
{\blue To find the polynomial order of the regret,
we use the non-linear least squares method to fit the curves to the
model $at^b+c$. The
resulting exponents $b$ of Exp3.R, D-UCB, Rexp3, SW-UCB, Exp3.S, CUSUM-UCB and PHT-UCB are
0.92, 0.89, 0.85, 0.84, 0.83, 0.72 and 0.69, respectively.} 
The regret of CUSUM-UCB and PHT-UCB shows the better sublinear
function of time compared to the other policies. Another
observation is that PHT-UCB performs better than CUSUM-UCB,
{\blue although we could not find a regret upper bound for
PHT-UCB}. The reason behind is that the PHT test is more stable and
reliable (due to the updated estimation $\hat{y}_k$) in the switching
environment.


\subsection{Yahoo! Dataset}
\begin{figure}[t]
\centering
\begin{subfigure}[b]{0.45\textwidth}
    \includegraphics[width=\textwidth]{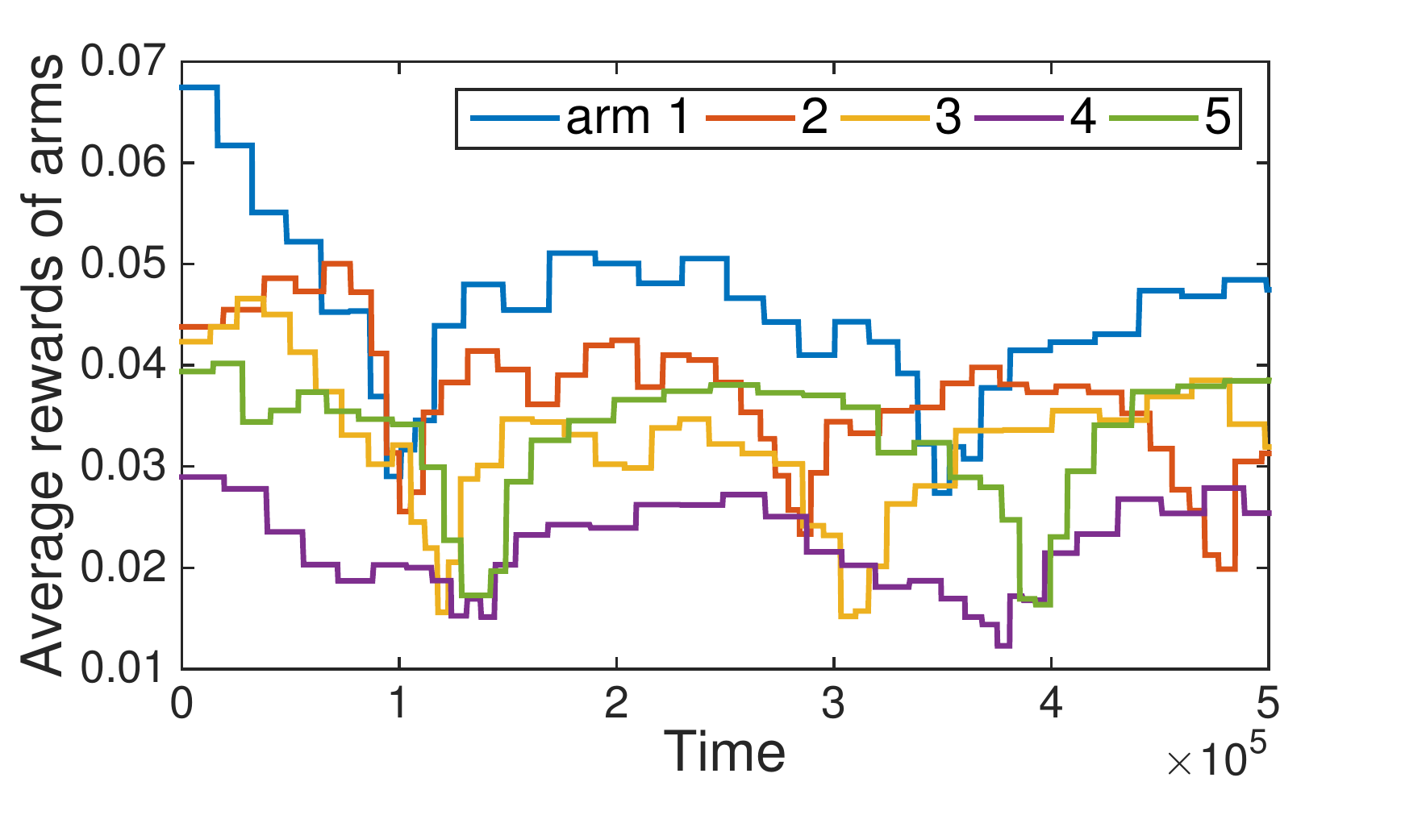}
     \caption{Ground truth}\label{subfig:truth}
\end{subfigure}
\begin{subfigure}[b]{0.45\textwidth}
    \includegraphics[width=\textwidth]{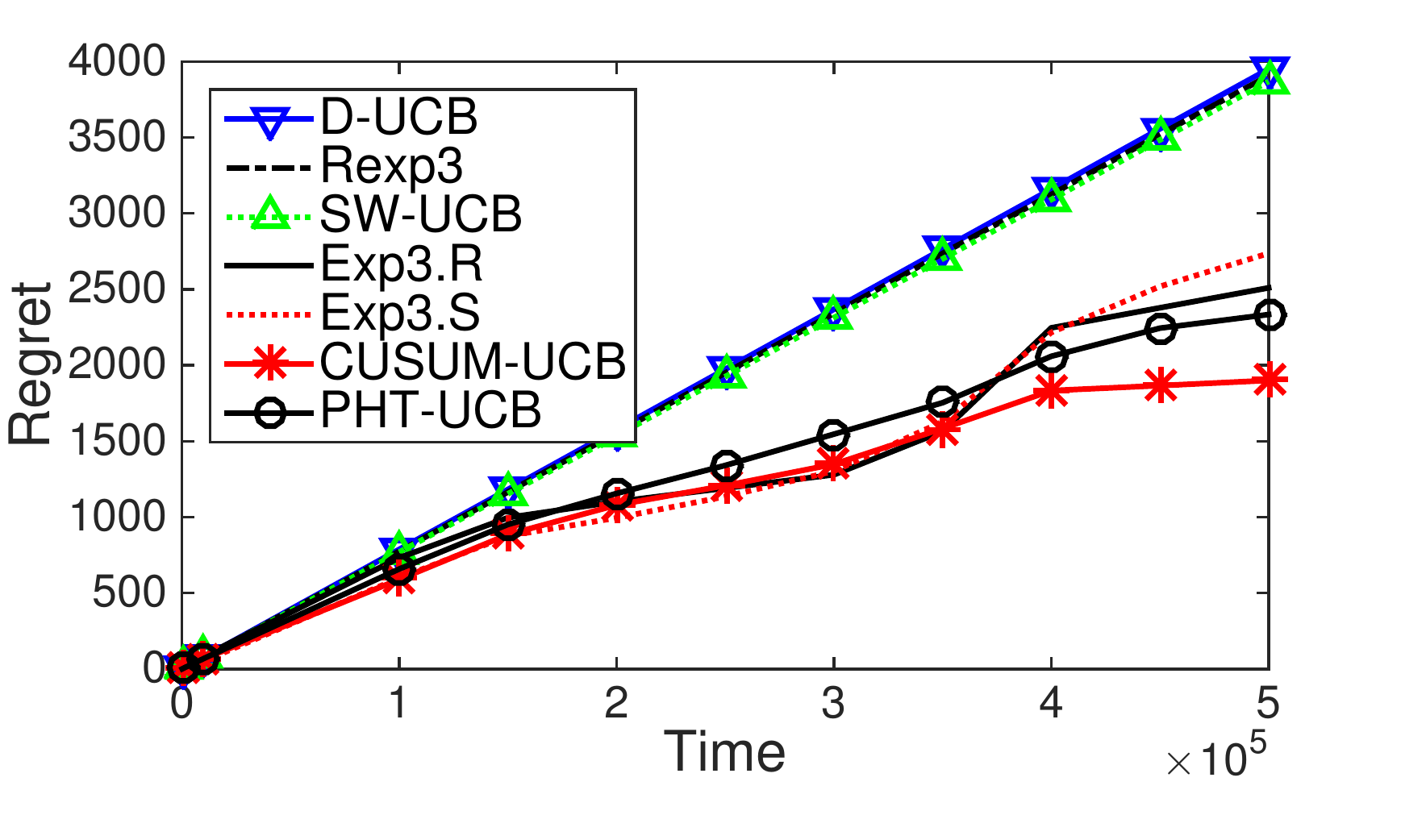}
     \caption{Regret}\label{subfig:result}
\end{subfigure}
\caption{Rewards and regret over the Yahoo! dataset with $K=5$}
\label{fig:yahoo}
\end{figure}

\begin{figure}[t]
\centering
\includegraphics[width=0.45\textwidth]{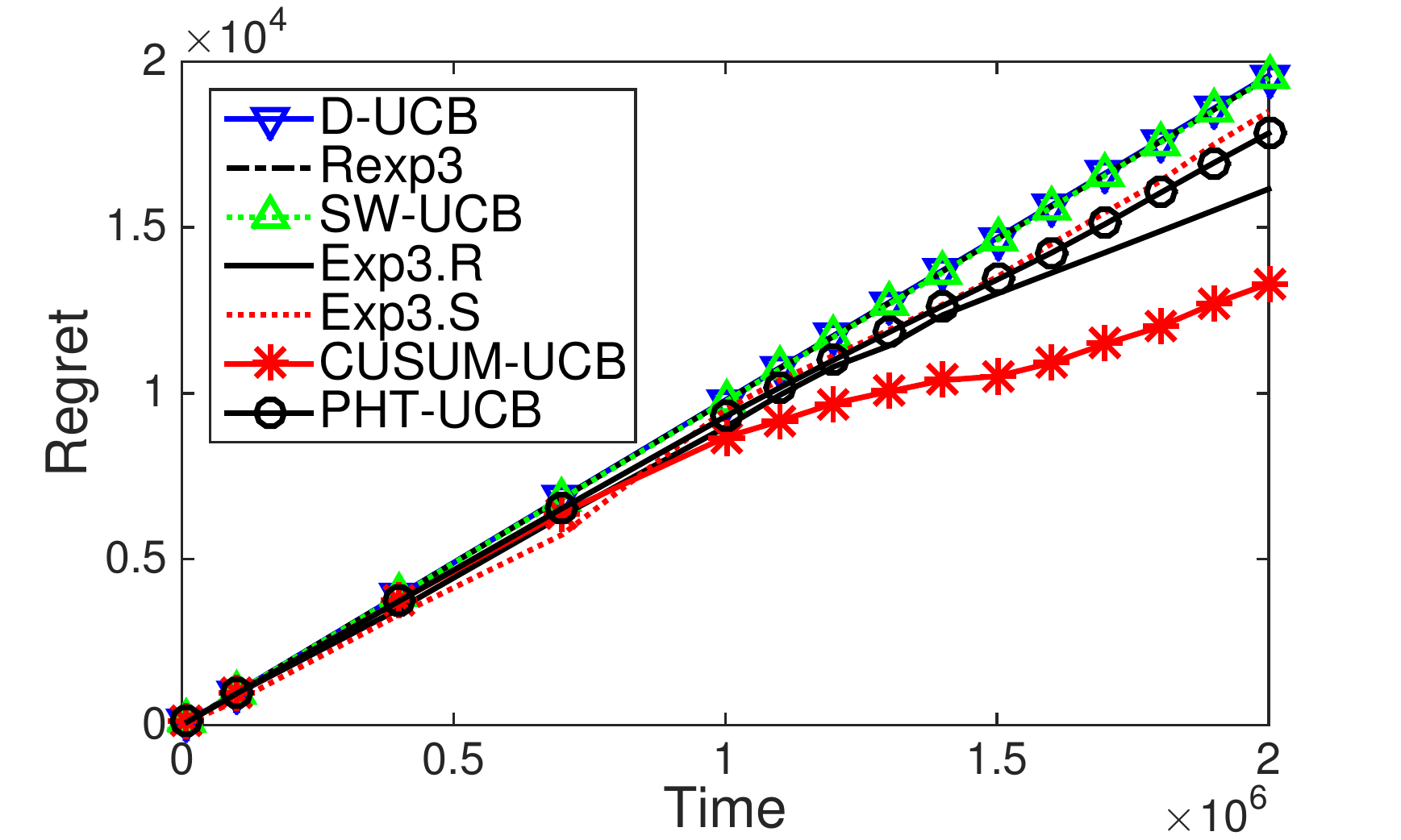}
\caption{Regret over the Yahoo! dataset with $K=100$}
\label{fig:yahoo_large}
\end{figure}
{\bf Yahoo! Experiment 1 ($K=5$).}
Yahoo! has published a benchmark dataset for the evaluation of bandit
algorithms~\cite{YahooWebScope}. The dataset is the user click
log for news articles displayed on the Yahoo! Front
Page~\cite{li2011unbiased}. Given the arrival of a user, the goal is
to select an article to present to the user, in order to maximize the
expected click-through rate,
{\blue where the reward is a binary value for user click.}
 For the purpose of our experiment, we
randomly select the set of 5 articles (i.e., $K=5$) from a list of
100 permutations of possible articles which overlapped in time the
most. To recover the ground truth of the expected click-through rates
of the articles, we take the same approach
as in~\citeauthor{mellor2013thompson}~\shortcite{mellor2013thompson}, where the click-through rates were estimated
from the dataset by taking the mean of an article's click-through rate
every 5000 time ticks (the length of a time tick is
about one second), which is shown in Figure~\ref{subfig:truth}.

The regret curves are shown in Figure~\ref{subfig:result}. 
{\blue We again fit the curves to the model $at^b+c$. The resulting
exponents $b$ of D-UCB, Rexp3, SW-UCB, Exp3.R, Exp3.S, CUSUM-UCB and PHT-UCB are 1, 1, 1, 0.81, 0.85, 0.69 and 0.79, respectively.}
The passively adaptive policies, D-UCB, SW-UCB and Rexp3, receive
a linear regret {\blue for most of the time.}
CUSUM-UCB and PHT-UCB achieve much better
performance and {\blue show} sublinear regret, because of their
active adaptation to changes.
Another observation is that CUSUM-UCB outperforms PHT-UCB. 
{\red The reason
behind is that the Yahoo! dataset has more {\blue frequent} breakpoints than the switching
environment {\blue (i.e., high $\gamma_T$)}.}
Thus, the estimation $\hat{y}_k$ in PHT test may drift away
before PHT detects the change, {\blue which in turn results in
  more detection misses and the higher regret.}

{\green
\noindent{\bf Yahoo! Experiment 2 ($K=100$).}
We repeat the above experiment with $K=100$. The regret curves are shown in Figure \ref{fig:yahoo_large}. We again fit the curves to the model $at^b+c$. The resulting exponents $b$ of D-UCB, Rexp3, SW-UCB, Exp3.R, Exp3.S, CUSUM-UCB and PHT-UCB are 1, 1, 1, 0.88, 0.9, 0.85 and 0.9, respectively. The passively adaptive policies, D-UCB, SW-UCB and Rexp3, receive
a linear regret for most of the time. CUSUM-UCB and PHT-UCB show robust performance in this larger scale experiment.
}
\section{Conclusion}\label{sec:conclusion}
We propose a change-detection based framework for multi-armed bandit
problems in the non-stationary setting. We study a class of
change-detection based policies, CD-UCB, and provide a general regret
upper bound given the performance of change detection algorithms. We
then develop CUSUM-UCB and PHT-UCB, that actively react to the environment by
detecting breakpoints. 
We analytically show that the regret of CUSUM-UCB is
$O(\sqrt{T\gamma_T\log{\frac{T}{\gamma_T}}})$, which is lower than the
regret bound of existing policies for the non-stationary setting.
To the best of our knowledge, this is the first regret bound for
actively adaptive UCB policies.
Finally, we demonstrate that CUSUM-UCB outperforms existing policies
via extensive experiments over arbitrary Bernoulli rewards and the
real world dataset of webpage click-through rates.

\section*{Acknowledgment}
This work has been supported in part by grants  from the Army Research Office  W911NF-14-1-0368 W911NF-15-1-0277, and MURI W911NF-12-1-0385, DTRA grant HDTRA1-14-1-0058, and NSF grant CNS-1719371.

\bibliographystyle{aaai}
\bibliography{refs}

\begin{thebibliography}{}

\bibitem[\protect\citeauthoryear{Agrawal and Goyal}{2012}]{agrawal2012analysis}
Agrawal, S., and Goyal, N.
\newblock 2012.
\newblock Analysis of thompson sampling for the multi-armed bandit problem.
\newblock In {\em COLT},  39.1--39.26.

\bibitem[\protect\citeauthoryear{Alaya-Feki, Moulines, and
  LeCornec}{2008}]{alaya2008dynamic}
Alaya-Feki, A. B.~H.; Moulines, E.; and LeCornec, A.
\newblock 2008.
\newblock Dynamic spectrum access with non-stationary multi-armed bandit.
\newblock In {\em 2008 IEEE 9th Workshop on Signal Processing Advances in
  Wireless Communications},  416--420.
\newblock IEEE.

\bibitem[\protect\citeauthoryear{Allesiardo and
  F{\'e}raud}{2015}]{allesiardo2015exp3}
Allesiardo, R., and F{\'e}raud, R.
\newblock 2015.
\newblock Exp3 with drift detection for the switching bandit problem.
\newblock In {\em Data Science and Advanced Analytics (DSAA), 2015. 36678 2015.
  IEEE International Conference on},  1--7.
\newblock IEEE.

\bibitem[\protect\citeauthoryear{Auer \bgroup et al\mbox.\egroup
  }{2002}]{doi:10.1137/S0097539701398375}
Auer, P.; Cesa-Bianchi, N.; Freund, Y.; and Schapire, R.~E.
\newblock 2002.
\newblock The nonstochastic multiarmed bandit problem.
\newblock {\em SIAM Journal on Computing} 32(1):48--77.

\bibitem[\protect\citeauthoryear{Auer, Cesa-Bianchi, and
  Fischer}{2002}]{auer2002finite}
Auer, P.; Cesa-Bianchi, N.; and Fischer, P.
\newblock 2002.
\newblock Finite-time analysis of the multiarmed bandit problem.
\newblock {\em Machine learning} 47(2-3):235--256.

\bibitem[\protect\citeauthoryear{Basseville and
  Nikiforov}{1993}]{basseville1993detection}
Basseville, M., and Nikiforov, I.~V.
\newblock 1993.
\newblock {\em Detection of abrupt changes: theory and application}, volume
  104.
\newblock Prentice Hall Englewood Cliffs.

\bibitem[\protect\citeauthoryear{Besbes, Gur, and
  Zeevi}{2014}]{besbes2014stochastic}
Besbes, O.; Gur, Y.; and Zeevi, A.
\newblock 2014.
\newblock Stochastic multi-armed-bandit problem with non-stationary rewards.
\newblock In {\em Advances in neural information processing systems},
  199--207.

\bibitem[\protect\citeauthoryear{Bubeck and
  Cesa-Bianchi}{2012}]{bubeck2012regret}
Bubeck, S., and Cesa-Bianchi, N.
\newblock 2012.
\newblock Regret analysis of stochastic and nonstochastic multi-armed bandit
  problems.
\newblock {\em arXiv preprint arXiv:1204.5721}.

\bibitem[\protect\citeauthoryear{Buccapatnam \bgroup et al\mbox.\egroup
  }{2017}]{buccapatnam2017reward}
Buccapatnam, S.; Liu, F.; Eryilmaz, A.; and Shroff, N.~B.
\newblock 2017.
\newblock Reward maximization under uncertainty: Leveraging side-observations
  on networks.
\newblock {\em arXiv preprint arXiv:1704.07943}.

\bibitem[\protect\citeauthoryear{Cesa-Bianchi and
  Lugosi}{2006}]{cesa2006prediction}
Cesa-Bianchi, N., and Lugosi, G.
\newblock 2006.
\newblock {\em Prediction, learning, and games}.
\newblock Cambridge university press.

\bibitem[\protect\citeauthoryear{Gallager}{2012}]{gallager2012discrete}
Gallager, R.~G.
\newblock 2012.
\newblock {\em Discrete stochastic processes}, volume 321.
\newblock Springer Science \& Business Media.

\bibitem[\protect\citeauthoryear{Garivier and
  Moulines}{2008}]{garivier2008upper}
Garivier, A., and Moulines, E.
\newblock 2008.
\newblock On upper-confidence bound policies for non-stationary bandit
  problems.
\newblock {\em arXiv preprint arXiv:0805.3415}.

\bibitem[\protect\citeauthoryear{Hartland \bgroup et al\mbox.\egroup
  }{2007}]{hartland2007change}
Hartland, C.; Baskiotis, N.; Gelly, S.; Sebag, M.; and Teytaud, O.
\newblock 2007.
\newblock Change point detection and meta-bandits for online learning in
  dynamic environments.
\newblock {\em CAp}  237--250.

\bibitem[\protect\citeauthoryear{Hinkley}{1971}]{hinkley1971inference}
Hinkley, D.~V.
\newblock 1971.
\newblock Inference about the change-point from cumulative sum tests.
\newblock {\em Biometrika}  509--523.

\bibitem[\protect\citeauthoryear{Kaufmann, Korda, and
  Munos}{2012}]{kaufmann2012thompson}
Kaufmann, E.; Korda, N.; and Munos, R.
\newblock 2012.
\newblock Thompson sampling: An asymptotically optimal finite-time analysis.
\newblock In {\em International Conference on Algorithmic Learning Theory},
  199--213.
\newblock Springer.

\bibitem[\protect\citeauthoryear{Khan}{1981}]{khan1981note}
Khan, R.~A.
\newblock 1981.
\newblock A note on page's two-sided cumulative sum procedure.
\newblock {\em Biometrika}  717--719.

\bibitem[\protect\citeauthoryear{Kocsis and
  Szepesv{\'a}ri}{2006}]{kocsis2006discounted}
Kocsis, L., and Szepesv{\'a}ri, C.
\newblock 2006.
\newblock Discounted ucb.
\newblock In {\em 2nd PASCAL Challenges Workshop},  784--791.

\bibitem[\protect\citeauthoryear{Lai and Robbins}{1985}]{lai1985asymptotically}
Lai, T.~L., and Robbins, H.
\newblock 1985.
\newblock Asymptotically efficient adaptive allocation rules.
\newblock {\em Advances in applied mathematics} 6(1):4--22.

\bibitem[\protect\citeauthoryear{Li \bgroup et al\mbox.\egroup
  }{2011}]{li2011unbiased}
Li, L.; Chu, W.; Langford, J.; and Wang, X.
\newblock 2011.
\newblock Unbiased offline evaluation of contextual-bandit-based news article
  recommendation algorithms.
\newblock In {\em Proceedings of the fourth ACM international conference on Web
  search and data mining},  297--306.
\newblock ACM.

\bibitem[\protect\citeauthoryear{Li, Karatzoglou, and
  Gentile}{2016}]{li2016collaborative}
Li, S.; Karatzoglou, A.; and Gentile, C.
\newblock 2016.
\newblock Collaborative filtering bandits.
\newblock In {\em Proceedings of the 39th International ACM SIGIR conference on
  Research and Development in Information Retrieval},  539--548.
\newblock ACM.

\bibitem[\protect\citeauthoryear{Lorden}{1971}]{lorden1971procedures}
Lorden, G.
\newblock 1971.
\newblock Procedures for reacting to a change in distribution.
\newblock {\em The Annals of Mathematical Statistics}  1897--1908.

\bibitem[\protect\citeauthoryear{Mellor and Shapiro}{2013}]{mellor2013thompson}
Mellor, J., and Shapiro, J.
\newblock 2013.
\newblock Thompson sampling in switching environments with bayesian online
  change detection.
\newblock In {\em Proceedings of the Sixteenth International Conference on
  Artificial Intelligence and Statistics},  442--450.

\bibitem[\protect\citeauthoryear{Page}{1954}]{page1954continuous}
Page, E.~S.
\newblock 1954.
\newblock Continuous inspection schemes.
\newblock {\em Biometrika} 41(1/2):100--115.

\bibitem[\protect\citeauthoryear{Pollard}{1984}]{pollard1984convergence}
Pollard, D.
\newblock 1984.
\newblock {\em Convergence of Stochastic Processes}.
\newblock Springer.

\bibitem[\protect\citeauthoryear{Srivastava, Reverdy, and
  Leonard}{2014}]{srivastava2014surveillance}
Srivastava, V.; Reverdy, P.; and Leonard, N.~E.
\newblock 2014.
\newblock Surveillance in an abruptly changing world via multiarmed bandits.
\newblock In {\em Decision and Control (CDC), 2014 IEEE 53rd Annual Conference
  on},  692--697.
\newblock IEEE.

\bibitem[\protect\citeauthoryear{Sutton and
  Barto}{1998}]{sutton1998reinforcement}
Sutton, R.~S., and Barto, A.~G.
\newblock 1998.
\newblock {\em Reinforcement learning: An introduction}, volume~1.
\newblock MIT press Cambridge.

\bibitem[\protect\citeauthoryear{Thompson}{1933}]{thompson1933likelihood}
Thompson, W.~R.
\newblock 1933.
\newblock On the likelihood that one unknown probability exceeds another in
  view of the evidence of two samples.
\newblock {\em Biometrika} 25(3/4):285--294.

\bibitem[\protect\citeauthoryear{Wei, Hong, and Lu}{2016}]{wei2016tracking}
Wei, C.-Y.; Hong, Y.-T.; and Lu, C.-J.
\newblock 2016.
\newblock Tracking the best expert in non-stationary stochastic environments.
\newblock In {\em Advances In Neural Information Processing Systems},
  3972--3980.

\bibitem[\protect\citeauthoryear{Yahoo!}{}]{YahooWebScope}
Yahoo!
\newblock Webscope program.
\newblock
  \url{http://webscope.sandbox.yahoo.com/catalog.php?datatype=r&did=49}.
\newblock {[Online; accessed 18-Oct-2016]}.

\bibitem[\protect\citeauthoryear{Yu and Mannor}{2009}]{yu2009piecewise}
Yu, J.~Y., and Mannor, S.
\newblock 2009.
\newblock Piecewise-stationary bandit problems with side observations.
\newblock In {\em Proceedings of the 26th Annual International Conference on
  Machine Learning},  1177--1184.
\newblock ACM.

\end{thebibliography}
\newpage
\appendix
\section{Lemma List}
We will make use of the following standard facts.

\begin{lemma}\label{lem:Chernoff}
\emph{(Chernoff-Hoeffding bound)} Let $Y_1,\ldots,Y_n$ be random variables with common range $[0,1]$ and such that $\mathbb{E}[Y_t|Y_1,\ldots,Y_{t-1}]=u$. Let $S_n=Y_1+\cdots+Y_n$. Then for all $a\geq0$
\begin{align}
\mathbb{P}\{S_n\geq nu+a\}&\leq e^{-2a^2/n},\\
\mathbb{P}\{S_n\leq nu-a\}&\leq e^{-2a^2/n}.
\end{align}
\end{lemma}
Lemma~\ref{lem:Chernoff} states the well known Chernoff-Hoeffding inequalities, proof of which is referred to~\cite{pollard1984convergence}. The next two lemmas are used in the proof of Theorem~\ref{thm:CUSUM}.

\begin{lemma}\label{lem:Wald}
\emph{(Wald's identity)} Let $\{Y_k;k\geq 1\}$ be independent and identically distributed, and let $\Lambda(r)=\log\{\mathbb{E}[e^{rY_1}]\}$. Let $J(Y)$ be the interval of $r$ over which $\Lambda(r)$ exists. For each $n\geq1$, let $S_n=Y_1+\cdots+Y_n$. Let $a<0$ and $b>0$ be arbitrary, and let $L$ be the smallest $n$ for which either $S_n\geq b$ or $S_n\leq a$. Then for each $r\in J(Y)$,
\begin{equation}\label{eqn:Wald}
\mathbb{E}\left[\exp\left(rS_L-L\Lambda(r)\right)\right]=1.
\end{equation}
\end{lemma}
Proof of Lemma~\ref{lem:Wald} is referred to~\cite{gallager2012discrete}.

\begin{lemma}\label{lem:khan}
\emph{(Two-sided CUSUM)} Let $\eta_1,\eta_2,\ldots$ be independent random variables. For $\epsilon\geq0$, define $s_k^+=\eta_k-\epsilon$ and $s_k^-=-\eta_k-\epsilon$. Let $g^+_k=\max(0,g^+_{k-1}+s_k^+)$ and $g^-_k=\max(0,g^-_{k-1}+s_k^-)$. For $h>0$, define $H^+=\inf\{k:g_k^+\geq h\}$ and $H^-=\inf\{k:g_k^-\geq h\}$. Let $H=\min(H^+,H^-)$. Then, we have that
\begin{equation}
\frac{1}{\mathbb{E}[H]}=\frac{1}{\mathbb{E}[H^+]}+\frac{1}{\mathbb{E}[H^-]}.
\end{equation}
\end{lemma}
Note that the two-sided CUSUM algorithm (that signals a change at $H$) defined in Lemma~\ref{lem:khan} is called symmetric. Proof of Lemma~\ref{lem:khan} is referred to~\cite{khan1981note}

\section{Proof of Theorem~\ref{thm:CD-UCB}}
\begin{proof}
Recall that $i_t$ is the arm with the best UCB index at time $t$, i.e., $i_t=\arg\max_{i\in\mathcal{K}}\left(\bar{X}_t(i)+C_t(i)\right)$. In addition $r_t$ is the random arm sampled from uniform distribution.
At each time $t$, if the arm $i$ is played, then the CD-UCB algorithm is either sampling a random arm ($r_t=i$) or playing the arm with the best UCB index ($i_t=i$).
So, the probability that arm $i$ is chosen at time $t$ and arm $i$ is not the best arm is
\begin{align}\nonumber
\mathbb{P}\{I_t=i\neq i^*_t\}&\leq \alpha\mathbb{P}\{r_t=i\}+(1-\alpha)\mathbb{P}\{i_t=i\neq i^*_t\}\\
&\leq \alpha/K+(1-\alpha)\mathbb{P}\{i_t=i\neq i^*_t\}.
\end{align}
By the definition of $\tilde{N}_T(i)$ in equation (\ref{def:nti}), we have that
\begin{align}
\mathbb{E}[\tilde{N}_T(i)]&\leq \sum_{t=1}^T\mathbb{P}\{I_t=i\neq i^*_t\}\\
&\leq \sum_{t=1}^T\left(\alpha/K+(1-\alpha)\mathbb{P}\{i_t=i\neq i^*_t\}\right)\\
&\leq \alpha T/K +
\underbrace{\sum_{t=1}^T\mathbb{P}\{i_t=i\neq i^*_t\}}_{(a)}.
\label{regretnti}
\end{align}
Now, it remains to bound the second term of (\ref{regretnti}), denoted by $(a)$. Let $A_i$ be a constant defined as $A_i\triangleq\frac{4\xi\log T}{(\Delta_{\mu_T}(i))^2}$. Then we can decompose the event $\{i_t=i\neq i^*_t\}$ as
\begin{equation*}
\{i_t=i\neq i^*_t,N_t(i)<A_i\}\cup\{i_t=i\neq i^*_t,N_t(i)\geq A_i\}.
\end{equation*}
Consider an experiment of the CD-UCB over $T$ plays. Let $F_i$ be the number of false alarms up to time $T$ and $D_i^j$ be the detection delay of $j$-th breakpoint on arm $i$, where $j\leq \gamma_T$. Then, the total number of detection points, when the change detection algorithm CD$(\cdot,\cdot)$ signals an alarm on arm $i$, is upper-bounded by $\gamma_T+F_i$. Let $\tau_i(t)$ be the latest detection points (including false arms) up to time $t$. For each arm $i$, we define $\mathcal{T}_i$ as the set of time slots that no breakpoint occurs after $A_i$ time slots away from the detection points.
\begin{align}\nonumber
\mathcal{T}_i\triangleq\{&t\in\{1,\ldots,T\}:\mu_s(i)=\mu_t(i)\\
&\text{and } \tau_i(t)<s\leq t, t\geq \tau_i(t)+A_i\}.
\end{align}
Then, we can bound the term $(b)\triangleq\sum_{t=1}^T\mathbbm{1}_{\{i_t=i\neq i^*_t\}}$ by
\begin{align}\nonumber
(b)&=\sum_{t=1}^T\left\{\mathbbm{1}_{\{i_t=i\neq i^*_t,N_t(i)<A_i\}}+\mathbbm{1}_{\{i_t=i\neq i^*_t,N_t(i)\geq A_i\}}\right\}\\
&\leq (\gamma_T+F_i)A_i+\sum_{t=1}^T\mathbbm{1}_{\{i_t=i\neq i^*_t,N_t(i)\geq A_i\}}\\\nonumber
&\leq (\gamma_T+F_i)A_i+\sum_{k=1}^{\gamma_T}D_i^k+\sum_{t\in\mathcal{T}_i}\mathbbm{1}_{\{i_t=i\neq i^*_t,N_t(i)\geq A_i\}}.
\end{align}
For each $t\in\mathcal{T}_i$, the event $\{i_t=i\neq i^*_t,N_t(i)\geq A_i\}$ implies the following union of events
\begin{align}\nonumber
\{\bar{X}_t(i)&\geq \mu_t(i)+C_t(i)\}\cup\{\bar{X}_t(i^*_t)\leq\mu_t(i^*_t)-C_t(i^*_t)\}\\
&\cup\{\mu_t(i^*_t)-\mu_t(i)<2C_t(i), N_t(i)\geq A_i\}.
\label{events}
\end{align}
On the event $\{N_t(i)\geq A_i\}$, we have that
\begin{equation}
C_t(i)=\sqrt{\frac{\xi\log n_t}{N_t(i)}}\leq \sqrt{\frac{\xi\log T}{A_i}}=\frac{\Delta_{\mu_T}(i)}{2},
\end{equation}
by the definition of $A_i$. Thus, $2C_t(i)\leq \mu_t(i^*_t)-\mu_t(i)$ holds when $N_t(i)\geq A_i$. This implies that the last event of (\ref{events}) is impossible. So we have that $ \mathbbm{1}_{\{i_t=i\neq i^*_t,N_t(i)\geq A_i\}}$ is at most
\begin{equation}
\mathbbm{1}_{\{\bar{X}_t(i)\geq \mu_t(i)+C_t(i)\}}+\mathbbm{1}_{\{\bar{X}_t(i^*_t)\leq\mu_t(i^*_t)-C_t(i^*_t)\}}
\end{equation}
By the Lemma~\ref{lem:Chernoff}, we have that
\begin{align}
\mathbb{P}{\{\bar{X}_t(i)\geq \mu_t(i)+C_t(i)\}}&\leq n_t^{-2\xi}\\
\mathbb{P}{\{\bar{X}_t(i^*_t)\leq\mu_t(i^*_t)-C_t(i^*_t)\}}&\leq n_t^{-2\xi}
\end{align}
Let $l_0,l_1,\ldots$, be the length of intervals between successive detection points. Then, we have that $\sum_{m=0}^{\gamma_T+F_i}l_m=T$.
Let $\xi=1$, we have that
\begin{align}
\sum_{t\in\mathcal{T}_i}\mathbb{P}&\{i_t=i\neq i^*_t,N_t(i)\geq A_i\}\leq\sum_{t\in\mathcal{T}_i}2n_t^{-2}\\
&\leq\sum_{t=1}^T2n_t^{-2}\leq 2\sum_{m=0}^{\gamma_T+F_i}\sum_{s=1}^{l_m}s^{-2}\\
&\leq2\sum_{m=0}^{\gamma_T+F_i}\pi^2/6=(\gamma_T+F_i+1)\pi^2/3.
\end{align}
Let $\mathbb{E}[F]$ be the expected number of false alarms in time horizon $T$, and $\mathbb{E}[D]$ be the expected detection delay. Summing all the results, we have $(a)=\mathbb{E}[(b)]$ and
\begin{align}\nonumber
(a)\leq&(\gamma_T+\mathbb{E}[F])A_i+{\gamma_T}\mathbb{E}[D]\\\label{terma}
&+\mathbb{E}\left[\sum_{t\in\mathcal{T}_i}\mathbb{P}{\{i_t=i\neq i^*_t,N_t(i)\geq A_i\}}\right].\\\nonumber
\leq&(\gamma_T+\mathbb{E}[F])A_i+{\gamma_T}\mathbb{E}[D]+(\gamma_T+\mathbb{E}[F]+1)\pi^2/3.
\end{align}
Combining (\ref{terma}) and (\ref{regretnti}), we obtain (\ref{thm1}).
\end{proof}

\section{Proof of Proposition~\ref{prop:CUSUM}}
\begin{proof}
Let $y_1,y_2,\ldots$ be a sequence of independent random variables with bounded support $[0,1]$. Before (after) the change, the random variable $y_k$ follows a distribution with parameter $\theta_0$ ($\theta_1$). Let $u_0$ ($u_1$) denote the mean before (after) the change. Under the Assumption \ref{ass:detectability}, we have that $|u_0-u_1|\geq2\epsilon$.
Note that we use the first $M$ samples to estimate the mean before the change by $\hat{u}_0$. Then, $s_k^-$ and $s_k^+$ become a non-trivial function of observation $y_k$. We define the expected time excluding the first $M$ time slots until the alarm occurs as the average run length (ARL), which is consistent with the literature in change detection problems~\cite{basseville1993detection}. In particular, let $L(\theta)$ be the ARL function defined as
\begin{equation}
L(\theta)=\mathbb{E}_{\theta}\left[\inf\{t:g^+_{t+M}>h \text{~or~} g^-_{t+M}>h\}\right].
\end{equation}
It is clear that Algorithm~\ref{alg:CUSUM} is a symmetric version of two-sided CUSUM-type algorithm~\cite{khan1981note}. Let $s_k^+$ and $g_k^+$ ($s_k^-$ and $g_k^-$) be the upper (lower) side CUSUM random walk. Then, we define the ARL function for lower and upper side CUSUM as
\begin{align}
L^-(\theta)&=\mathbb{E}_{\theta}\left[\inf\{t:g^-_{t+M}>h\}\right],\\
L^+(\theta)&=\mathbb{E}_{\theta}\left[\inf\{t:g^+_{t+M}>h\}\right].
\end{align}
By Lemma~\ref{lem:khan}, we have that 
\begin{equation}\label{eqn:khan}
\frac{1}{L(\theta)}=\frac{1}{L^-(\theta)}+\frac{1}{L^+(\theta)}
\end{equation}

The ARL function $L(\theta)$ characterizes the detection delay and false alarm performances of our CUSUM algorithm. In particular, $L(\theta_0)$ is the mean detection delay, and $L(\theta_1)$ is the mean time between false alarms. Thus, $\mathbb{E}[D]=L(\theta_0)$ and $\mathbb{E}[F]=T/L(\theta_1)$. By (\ref{eqn:khan}), it remains to calculate $L^-(\theta)$ and $L^+(\theta)$. In the following, we show the results for $L^-(\theta)$ and obtain the results for $L^+(\theta)$ by symmetry.

Let $S^-_n=s^-_{1+M}+\ldots+s^-_{n+M}$ be a random walk. Let $H^-$ be the smallest $n$ such that $S^-_n$ crosses the bounded interval $[0,h]$, i.e., $H^-=\inf\{S^-_n<0 \text{ or } S^-_n>h\}$. The procedure that monitors the stopping time that $S^-_n$ crosses the boundary is also called a sequential probability ratio test (SPRT). Then the lower side CUSUM can be viewed as a repeated SPRT such that CUSUM restarts the SPRT once $S^-_n$ crosses the $0$ boundary (i.e., $S^-_n<0$) and outputs an alarm the first time that $S^-_n$ crosses the $h$ boundary (i.e., $S^-_n>h$)~\cite{page1954continuous}. Let $\mathbb{P}_\theta\{0\}\triangleq\mathbb{P}_\theta\{S^-_{H^-}<0\}$ be the probability that SPRT ends up with $S^-_{H^-}<0$. Let $G$ be the number of SPRT tests that CUSUM runs until an alarm. Then $G-1$ is a geometrically distributed random variable with parameter $\mathbb{P}_\theta\{0\}$. Hence, we have that
\begin{align}\nonumber
L^-(\theta)&=\mathbb{E}_\theta[H^-|S^-_{H^-}\!<\! 0]\mathbb{E}_\theta[G \!-\! 1] + \mathbb{E}_\theta[H^-|S^-_{H^-}\!>\! h]\\
\label{arllong}
&=\frac{\mathbb{E}_\theta[H^-|S^-_{H^-}<0]\mathbb{P}_\theta\{0\}}{1-\mathbb{P}_\theta\{0\}} + \mathbb{E}_\theta[H^-|S^-_{H^-}>h]\\
\label{arlshort}
&=\frac{\mathbb{E}_\theta[H^-]}{1-\mathbb{P}_\theta\{0\}}.
\end{align}
Let $\Lambda^-_\theta(r)\triangleq \log\{\mathbb{E}_\theta[e^{rs^-_{M+1}}]\}$. Then, by Lemma~\ref{lem:Wald} we have 
\begin{equation}\label{eqn:wald}
\mathbb{E}_\theta\left[\exp\left(rS_{H^-}-H^-\Lambda^-_\theta(r)\right)\right]=1.
\end{equation}
We use the two implications from the Wald's identity
$(\ref{eqn:wald})$. For simplicity, we assume that $k\geq M$ when we consider the expectation of $s^-_k$ and $s^+_k$. First, taking the derivative of both sides and letting $r=0$, we have that
\begin{equation}\label{eqn:waldeqn}
\mathbb{E}_\theta[S_{H^-}]=\mathbb{E}_\theta[H^-]\mathbb{E}_\theta[s^-_k].
\end{equation}
Second, letting $r=r^-(\theta)$ such that $\Lambda^-_\theta(r^-(\theta))=0$ and $r^-(\theta)\neq0$, we have that
\begin{equation}\label{eqn:waldeqn2}
\mathbb{E}_\theta[\exp{(r^-(\theta)S_{H^-})}]=1.
\end{equation}

By the Assumption~\ref{ass:piecewise}, $\hat{u}_0$ is the average of $M$ samples from distribution under $\theta_0$. By Lemma~\ref{lem:Chernoff}, we have that
\begin{equation}
\mathbb{P}\{|\hat{u}_0-u_0|>\epsilon\}\leq2e^{-2\epsilon^2M}.
\end{equation}
Now, we classify the possible scenarios into four different cases depending on $\hat{u}_0, u_0$ and $u_1$, under which we can derive the upper bound of the mean detection delay or the mean time between false alarms.

\emph{Case 1:} $|\hat{u}_0-u_0|<\epsilon$ and $u_1<u_0$

Under $\theta_1$, we show the upper bound of $L^-(\theta_1)$ (mean detection delay of lower side CUSUM). Note that $\mathbb{E}_{\theta_1}[s^-_k]=\hat{u}_0-u_1-\epsilon>0$. Then, by (\ref{arlshort}) and (\ref{eqn:waldeqn})
we have that
\begin{align}
L^-(\theta_1)=&\frac{\mathbb{E}_{\theta_1}[H^-]}{1-\mathbb{P}_{\theta_1}\{0\}}=\frac{\mathbb{E}_{\theta_1}[S_{H^-}]}{\mathbb{E}_{\theta_1}[s^-_k](1-\mathbb{P}_{\theta_1}\{0\})}\\\nonumber
=&\frac{\mathbb{E}_{\theta_1}[S_{H^-}|S_{H^-}> h](1-\mathbb{P}_{\theta_1}\{0\})}{\mathbb{E}_{\theta_1}[s^-_k](1-\mathbb{P}_{\theta_1}\{0\})}\\
&+\frac{\mathbb{E}_{\theta_1}[S_{H^-}|S_{H^-}<0]\mathbb{P}_{\theta_1}\{0\}}{\mathbb{E}_{\theta_1}[s^-_k](1-\mathbb{P}_{\theta_1}\{0\})}\\
\leq&\frac{\mathbb{E}_{\theta_1}[S_{H^-}|S_{H^-}> h]}{\mathbb{E}_{\theta_1}[s^-_k]}\\
\leq&\frac{h+1}{\mathbb{E}_{\theta_1}[s^-_k]}.
\end{align}

Under $\theta_0$, we show the lower bound of $L^-(\theta_0)$ (mean time between false alarms of lower side CUSUM). Note that $\mathbb{E}_{\theta_0}[s^-_k]=\hat{u}_0-u_0-\epsilon<0$, which implies that $r^-(\theta_0)>0$. By (\ref{eqn:waldeqn2}), we have that
\begin{align}
1=&\mathbb{E}_{\theta_0}[\exp{(r^-(\theta_0)S_{H^-})}]\\
=&\mathbb{P}_{\theta_0}\{0\}\mathbb{E}_{\theta_0}[\exp{(r^-(\theta_0)S_{H^-})}|S_{H^-}<0]\\\nonumber
&+(1-\mathbb{P}_{\theta_0}\{0\})\mathbb{E}_{\theta_0}[\exp{(r^-(\theta_0)S_{H^-})}|S_{H^-}>h]\\
\geq&(1-\mathbb{P}_{\theta_0}\{0\})\mathbb{E}_{\theta_0}[\exp{(r^-(\theta_0)S_{H^-})}|S_{H^-}>h]\\
\geq&(1-\mathbb{P}_{\theta_0}\{0\})\exp{(r^-(\theta_0)h)}
\end{align}
Note that $\mathbb{E}_{\theta_0}[H^-]\geq1$. Hence, we have that
\begin{align}
L^-(\theta_0)=\frac{\mathbb{E}_{\theta_0}[H^-]}{1-\mathbb{P}_{\theta_0}\{0\}}\geq\frac{1}{1-\mathbb{P}_{\theta_0}\{0\}}\geq\exp{(r^-(\theta_0)h)}.
\end{align}

Under $\theta_0$, we obtain the lower bound of $L^+(\theta_0)$ (mean time between false alarms of upper side CUSUM) with the similar arguments. In particular, $\mathbb{E}_{\theta_0}[s^+_k]=u_0-\hat{u}_0-\epsilon<0$, which implies that $r^+(\theta_0)>0$. Then we have that
\begin{equation}
L^+(\theta_0)\geq\exp{(r^+(\theta_0)h)}.
\end{equation}
Let $r(\theta_0)=\min(r^+(\theta_0),r^-(\theta_0))$. By (\ref{eqn:khan}), we have that
\begin{align}
L(\theta_1)\leq& L^-(\theta_1)\leq\frac{h+1}{\mathbb{E}_{\theta_1}[s^-_k]},\\
L(\theta_0)\geq&\frac{\exp{(r(\theta_0)h)}}{2}.
\end{align}

\emph{Case 2:} $|\hat{u}_0-u_0|<\epsilon$ and $u_1>u_0$\\
Similarly, we obtain the results for upper side CUSUM as
\begin{align}
L^+(\theta_1)\leq&\frac{h+1}{\mathbb{E}_{\theta_1}[s^+_k]}\\
L^+(\theta_0)\geq&\exp{(r^+(\theta_0)h)}.
\end{align}
Under $\theta_0$, we check that $\mathbb{E}_{\theta_0}[s^-_k]=\hat{u}_0-u_0-\epsilon<0$. Hence, we have that
\begin{equation}
L^-(\theta_0)\geq\exp{(r^-(\theta_0)h)}.
\end{equation}
By (\ref{eqn:khan}), we have that
\begin{align}
L(\theta_1)\leq& L^+(\theta_1)\leq\frac{h+1}{\mathbb{E}_{\theta_1}[s^+_k]},\\
L(\theta_0)\geq&\frac{\exp{(r(\theta_0)h)}}{2}.
\end{align}

\emph{Case 3:} $\hat{u}_0-u_0>\epsilon$

When the estimate $\hat{u}_0$ is large ($\hat{u}_0-u_0>\epsilon$), $\mathbb{E}_{\theta_0}[s^-_k]=\hat{u}_0-u_0-\epsilon>0$. Then we have that 
\begin{equation}
L(\theta_0)\leq L^-(\theta_0)\leq \frac{h+1}{\mathbb{E}_{\theta_0}[s_k^-]}.
\end{equation}

\emph{Case 4:} $u_0-\hat{u}_0>\epsilon$ 

When the estimate $\hat{u}_0$ is small ($u_0-\hat{u}_0>\epsilon$), $\mathbb{E}_{\theta_0}[s^+_k]=u_0-\hat{u}_0-\epsilon>0$. Then we have that 
\begin{equation}
L(\theta_0)\leq L^+(\theta_0)\leq \frac{h+1}{\mathbb{E}_{\theta_0}[s_k^+]}.
\end{equation}
\end{proof}

\section{Proof of Theorem~\ref{thm:CUSUM}}
Note that the bounds for $L(\theta)$, $L^-(\theta)$ and $L^+(\theta)$ in the Proposition~\ref{prop:CUSUM} are the conditional expectations under $\theta$ given $\hat{u}_0$. By the law of total expectation, we can obtain the expected bound results by taking expectation over $\hat{u}_0$. Let $\mathbb{E}_{\hat{u}_0}$ denote the expectation over $\hat{u}_0$. We classify the possible scenarios into two cases, under which we take the conditional expectations and sum the total expectation finally.

\emph{Case 1:} $|\hat{u}_0-u_0|<\epsilon$

By the Proposition~\ref{prop:CUSUM}, given the condition $\hat{u}_0$ and $|\hat{u}_0-u_0|<\epsilon$, we have that
\begin{align}
L(\theta_1)&\leq \frac{h+1}{|u_1-\hat{u}_0|-\epsilon}\\
L^-(\theta_0)&\geq \exp(r^-(\theta_0)h)\\
L^+(\theta_0)&\geq \exp(r^+(\theta_0)h)
\end{align}
First, we consider the upper bound of $L(\theta_1)$ when $u_1<u_0$. Let $p_{Z}(\cdot)$ denote the probability density function (or probability mass function) of random variable $Z$. Then, we have that
\begin{align}
\mathbb{E}_{\hat{u}_0}&\left[\frac{h+1}{|u_1-\hat{u}_0|-\epsilon}\middle||\hat{u}_0-u_0|<\epsilon\right]\\
=&\left[\int_{u_0-\epsilon}^{u_0+\epsilon}\frac{h+1}{|u_1-u|-\epsilon}p_{\hat{u}_0}(u)du\right]/\mathbb{P}\{|\hat{u}_0-u_0|<\epsilon\}\\
=&\left[\int_{u_0-\epsilon}^{u_0+\epsilon}\frac{h+1}{u-u_1-\epsilon}p_{\hat{u}_0}(u)du\right]/\mathbb{P}\{|\hat{u}_0-u_0|<\epsilon\}
\end{align}
We define
$\delta_T(i)\triangleq\min\{|\mu_t(i)-\mu_{t+1}(i)|-2\epsilon:\mu_t(i)\neq
\mu_{t+1}(i), t\leq T\}$. Note that $\delta_T(i)$ is the minimum over a finite set of positive
real numbers (i.e., $\delta_T(i)>0$). Then,
$u_0-u_1-\epsilon>\delta_T(i)+\epsilon$. Therefore, we have 
\begin{align}
\int_{u_0-\epsilon}^{u_0+\epsilon}&\frac{1}{u-u_1-\epsilon}p_{\hat{u}_0}(u)du\\
\leq&\int_{-\epsilon}^{\epsilon}\frac{1}{u+u_0-u_1-\epsilon}du\\
\leq&\int_{-\epsilon}^{\epsilon}\frac{1}{u+\delta_T(i)+\epsilon}du\\
=&\log(1+2\epsilon/\delta_T(i)).
\end{align}
Hence, we have that
\begin{align}
\mathbb{E}_{\hat{u}_0}&\left[\frac{h+1}{|u_1-\hat{u}_0|-\epsilon}\middle||\hat{u}_0-u_0|<\epsilon\right]\\
\leq&\frac{(h+1)\log(1+2\epsilon/\delta_T(i))}{\mathbb{P}\{|\hat{u}_0-u_0|<\epsilon\}}
\end{align}
The same result holds for $u_1>u_0$. Note that $\delta_T(i)\geq\epsilon$ by the Assumption~\ref{ass:detectability}. Then, we have that
\begin{equation}\label{case1D}
\mathbb{E}_{\hat{u}_0}\left[\frac{h+1}{|u_1-\hat{u}_0|-\epsilon}\middle||\hat{u}_0-u_0|<\epsilon\right]\leq\frac{(h+1)\log(3)}{\mathbb{P}\{|\hat{u}_0-u_0|<\epsilon\}}
\end{equation}

Second, we consider the lower bound of $L^-(\theta_0)$ and $L^+(\theta_0)$. By the Assumption~\ref{ass:lambda}, the $y_k$ is a Bernoulli random variable. Then, we have that
\begin{align}
\Lambda_{\theta_0}^-(r)&=\log(e^{-r}u_0+1-u_0)+r(\hat{u}_0-\epsilon)\\
\Lambda_{\theta_0}^+(r)&=\log(e^{r}u_0+1-u_0)-r(\hat{u}_0+\epsilon)
\end{align}
Let $\hat{r}^-(\theta_0)$ ($\hat{r}^+(\theta_0)$) be the solution of $\frac{d}{dr}\Lambda^-_{\theta_0}(r)=0$ $\left(\frac{d}{dr}\Lambda^+_{\theta_0}(r)=0\right)$. Then, the convexity of $\Lambda_{\theta_0}^-(r)$ $\left(\Lambda_{\theta_0}^+(r)\right)$ implies that $r^-(\theta_0)>\hat{r}^-(\theta_0)$ $\left(r^-(\theta_0)>\hat{r}^-(\theta_0)\right)$. In addition, $\hat{r}^-(\theta_0)$ and $\hat{r}^+(\theta_0)$ satisfy
\begin{align}\label{rlower}
\exp(\hat{r}^-(\theta_0))&=\frac{u_0(\hat{u}_0-\epsilon)^{-1}-u_0}{1-u_0},\\\label{rupper}
\exp(\hat{r}^+(\theta_0))&=\frac{1-u_0}{u_0(\hat{u}_0+\epsilon)^{-1}-u_0}.
\end{align} 
Note that $\hat{r}^-(\theta_0)$ and $\hat{r}^+(\theta_0)$ always exist when $2\epsilon<u_0<1-2\epsilon$. One can scale the reward linearly without changing the problem by function $f(x)=(1-4\epsilon)x+2\epsilon$ so that the expected reward is within the interval $(2\epsilon,1-2\epsilon)$. Hence, we assume that $\hat{r}^-(\theta_0)$ and $\hat{r}^+(\theta_0)$ exist without loss of generality. We first derive a lower bound for (\ref{rlower}).
\begin{align}
&\mathbb{E}_{\hat{u}_0} \left[(\hat{u}_0-\epsilon)^{-1}\middle||\hat{u}_0-u_0|<\epsilon\right]\\
&=\frac{\int_{u_0-\epsilon}^{u_0+\epsilon}\frac{1}{u-\epsilon}p_{\hat{u}_0}(u)du}{\mathbb{P}\{|\hat{u}_0-u_0|<\epsilon\}}\\
&\geq \frac{\int_{u_0-\epsilon}^{u_0+\epsilon}\left(\frac{1}{u-\epsilon}-\frac{1}{u_0}\right)p_{\hat{u}_0}(u)du}{\mathbb{P}\{|\hat{u}_0-u_0|<\epsilon\}}+\frac{1}{u_0}\\
&\geq \frac{\int_{u_0-\epsilon}^{u_0}\left(\frac{1}{u-\epsilon}-\frac{1}{u_0}\right)p_{\hat{u}_0}(u)du}{\mathbb{P}\{|\hat{u}_0-u_0|<\epsilon\}}+\frac{1}{u_0}\\
&\geq \frac{\left(\frac{1}{u_0-\epsilon}-\frac{1}{u_0}\right)\int_{u_0-\epsilon}^{u_0}p_{\hat{u}_0}(u)du}{\mathbb{P}\{|\hat{u}_0-u_0|<\epsilon\}}+\frac{1}{u_0}\\
&\geq\frac{\epsilon}{u_0(u_0 \!-\!\epsilon)}\binom{M}{\lfloor u_0M\rfloor}u_0^{\lfloor u_0M\rfloor}(1\!-\! u_0)^{M-{\lfloor u_0M\rfloor}}+\frac{1}{u_0}\\
&\geq\frac{\epsilon}{u_0(u_0-\epsilon)}\binom{M}{\lfloor 2\epsilon M\rfloor}(2\epsilon)^M+\frac{1}{u_0}.
\end{align}
Hence, we have that
\begin{align}
\mathbb{E}_{\hat{u}_0}&\left[\exp(\hat{r}^-(\theta_0))\middle||\hat{u}_0-u_0|<\epsilon\right]\\
&\geq\frac{\epsilon}{(u_0-\epsilon)(1-u_0)}\binom{M}{\lfloor 2\epsilon M\rfloor}(2\epsilon)^M+1\\
&\geq\frac{4\epsilon}{(1-\epsilon)^2}\binom{M}{\lfloor 2\epsilon M\rfloor}(2\epsilon)^M+1.
\end{align}
Now, we derive a lower bound for (\ref{rupper}). Similarly, we have that
\begin{align}
\mathbb{E}_{\hat{u}_0}&\left[(\hat{u}_0+\epsilon)^{-1}\middle||\hat{u}_0-u_0|<\epsilon\right]\\
&=\frac{\int_{u_0-\epsilon}^{u_0+\epsilon}\frac{1}{u+\epsilon}p_{\hat{u}_0}(u)du}{\mathbb{P}\{|\hat{u}_0-u_0|<\epsilon\}}\\
&\leq \frac{\int_{u_0}^{u_0+\epsilon}\left(\frac{1}{u+\epsilon}-\frac{1}{u_0}\right)p_{\hat{u}_0}(u)du}{\mathbb{P}\{|\hat{u}_0-u_0|<\epsilon\}}+\frac{1}{u_0}\\
&\leq \frac{\left(\frac{1}{u_0+\epsilon}-\frac{1}{u_0}\right)\int_{u_0}^{u_0+\epsilon}p_{\hat{u}_0}(u)du}{\mathbb{P}\{|\hat{u}_0-u_0|<\epsilon\}}+\frac{1}{u_0}\\
&\leq \frac{-\epsilon}{u_0(u_0+\epsilon)}\binom{M}{\lceil 2\epsilon M\rceil}(2\epsilon)^M+\frac{1}{u_0}.
\end{align}
Hence, by the Jensen's inequality, we have that
\begin{align}
\mathbb{E}_{\hat{u}_0}&\left[\exp(\hat{r}^+(\theta_0))\middle||\hat{u}_0-u_0|<\epsilon\right]\\
&=\mathbb{E}_{\hat{u}_0}\left[\frac{1-u_0}{u_0(\hat{u}_0+\epsilon)^{-1}-u_0}\middle||\hat{u}_0-u_0|<\epsilon\right]\\
&\geq \frac{1-u_0}{\mathbb{E}_{\hat{u}_0}\left[{u_0(\hat{u}_0+\epsilon)^{-1}-u_0}\middle||\hat{u}_0-u_0|<\epsilon\right]}\\
&\geq \frac{1-u_0}{\frac{-\epsilon}{(u_0+\epsilon)}\binom{M}{\lceil 2\epsilon M\rceil}(2\epsilon)^M+{1}-{u_0}}\\
&=\frac{\epsilon\binom{M}{\lceil 2\epsilon M\rceil}(2\epsilon)^M}{(1-u_0)(u_0+\epsilon)-\epsilon\binom{M}{\lceil 2\epsilon M\rceil}(2\epsilon)^M}+1\\
&\geq \frac{\epsilon\binom{M}{\lceil 2\epsilon M\rceil}(2\epsilon)^M}{(1+\epsilon)^2/4-\epsilon\binom{M}{\lceil 2\epsilon M\rceil}(2\epsilon)^M}+1\\
&\geq \frac{4\epsilon}{(1+\epsilon)^2}\binom{M}{\lceil 2\epsilon M\rceil}(2\epsilon)^M+1
\end{align}
Let $C_1^-=\log\left(\frac{4\epsilon}{(1-\epsilon)^2}\binom{M}{\lfloor 2\epsilon M\rfloor}(2\epsilon)^M+1\right)$ and $C_1^+=\log\left(\frac{4\epsilon}{(1+\epsilon)^2}\binom{M}{\lceil 2\epsilon M\rceil}(2\epsilon)^M+1\right)$. Then, there exists a positive number $C_1\triangleq\min(C_1^-,C_1^+)$, which depends on $\epsilon$ and $M$, such that
\begin{align}
\mathbb{E}_{\hat{u}_0}&\left[\exp(r^-(\theta_0)h)\middle||\hat{u}_0-u_0|<\epsilon\right]\\
&\geq\mathbb{E}_{\hat{u}_0}\left[\exp(\hat{r}^-(\theta_0)h)\middle||\hat{u}_0-u_0|<\epsilon\right]\\
&\geq\left\{\mathbb{E}_{\hat{u}_0}\left[\exp(\hat{r}^-(\theta_0))\middle||\hat{u}_0-u_0|<\epsilon\right]\right\}^h\\\label{case1F1}
&\geq \exp(C_1h).
\end{align}
Similarly, we have that
\begin{equation}\label{case1F2}
\mathbb{E}_{\hat{u}_0}\left[\exp(r^+(\theta_0)h)\middle||\hat{u}_0-u_0|<\epsilon\right]\geq\exp(C_1h).
\end{equation}

\emph{Case 2:} $|\hat{u}_0-u_0|>\epsilon$

By Assumption \ref{ass:lambda}, the $\hat{u}_0$ is the average of $M$ Bernoulli random variables. In other words, $\hat{u}_0$ must be one of the points $\{0,1/M,\ldots,1\}$. Then there must be a gap between $\hat{u}_0$ and $u_0$. Let $\lambda_T(i)=\min\{(\mu_t(i)-\epsilon)-{\lfloor (\mu_t(i)-\epsilon)M\rfloor}/{M}, {\lceil (\mu_t(i)+\epsilon)M\rceil}/{M}-(\mu_t(i)+\epsilon):1\leq t \leq T\}$ be the minimal gap of arm $i$.\footnote{Note that $\lfloor\cdot\rfloor$ denotes the floor function and $\lceil\cdot\rceil$ denotes the ceiling function.} We define the minimal gap of all arms as $\lambda=\min_{i\in\mathcal{K}}\lambda_T(i)$.

By the Proposition~\ref{prop:CUSUM}, the average run length is at most
\begin{equation}
\frac{h+1}{|\hat{u}_0-u_0|-\epsilon}.
\end{equation}
Suppose that $\hat{u}_0>u_0+\epsilon$, then
\begin{align}
\mathbb{E}_{\hat{u}_0}&\left[\frac{h+1}{|\hat{u}_0-u_0|-\epsilon}\middle||\hat{u}_0-u_0|>\epsilon\right]\\
=&\mathbb{E}_{\hat{u}_0}\left[\frac{h+1}{\hat{u}_0-u_0-\epsilon}\middle||\hat{u}_0-u_0|>\epsilon\right]\\
\leq&\mathbb{E}_{\hat{u}_0}\left[\frac{h+1}{\lceil(u_0+\epsilon)M\rceil/M-u_0-\epsilon}\middle||\hat{u}_0-u_0|>\epsilon\right]\\\label{case2}
\leq&\frac{h+1}{\lambda}.
\end{align}
The same result holds when $\hat{u}_0<u_0-\epsilon$. 
Summing the results (\ref{case1D}) (\ref{case1F1}) (\ref{case1F2}) (\ref{case2}), we derive the upper bound for the mean detection delay and the lower bound for number of false alarms within the horizon $T$.
\begin{align}
\hspace{-0.05cm}\mathbb{E}[D]&\leq\mathbb{E}_{\hat{u}_0}\left[\frac{h+1}{|u_1 \!-\!\hat{u}_0|\!-\!\epsilon}\middle||\hat{u}_0 \!-\! u_0|<\epsilon\right]\mathbb{P}\{|\hat{u}_0 \!-\! u_0|\!<\!\epsilon\}\\
&\,\,+\mathbb{E}_{\hat{u}_0}\left[\frac{h+1}{|\hat{u}_0 \!-\! u_0|\!-\!\epsilon}\middle||\hat{u}_0 \!-\! u_0|>\epsilon\right]\mathbb{P}\{|\hat{u}_0 \!-\! u_0|\!>\!\epsilon\}\\
&\leq(h+1)\log(3)+2\exp(-2\epsilon^2M)\frac{h+1}{\lambda}.
\end{align}
Note that the mean time between false alarm is $\mathbb{E}_{\hat{u}_0}[L(\theta_0)]$. By the law of total expectation, we have that
\begin{equation}
\mathbb{E}_{\hat{u}_0}[L(\theta_0)]\geq\mathbb{E}_{\hat{u}_0}[L(\theta_0)||\hat{u}_0-u_0|<\epsilon]\mathbb{P}\{|\hat{u}_0-u_0|<\epsilon\}
\end{equation}
By the Lemma~\ref{lem:khan} and (\ref{case1F1}) (\ref{case1F2}), we have that
\begin{equation}
\mathbb{E}_{\hat{u}_0}[L(\theta_0)||\hat{u}_0-u_0|<\epsilon]\geq\frac{1}{2}\exp(C_1h)
\end{equation}
Hence, we have that
\begin{align}
\mathbb{E}[F]\leq&\frac{T}{\frac{1}{2}\exp(C_1h)\mathbb{P}\{|\hat{u}_0-u_0|<\epsilon\}}\\
\leq&\frac{2T}{\exp(C_1h)(1-2\exp(-2\epsilon^2M))}.
\end{align}

\section{Proof of Theorem~\ref{thm:CUSUM-UCB}}
The result follows the Theorem~\ref{thm:CD-UCB} and the
Theorem~\ref{thm:CUSUM}. Since the results of the
Theorem~\ref{thm:CUSUM} do not include the first $M$ time slots, we
additionally count the regret of $M$ for each detection point (alarm). The mean detection delay is at most scaled by dividing the sampling rate $\alpha/K$. 

\section{CUSUM-UCB policy}\label{app:cusumucb}
CUSUM-UCB policy is a CD-UCB policy with CUSUM as a change detection algorithm. In particular, it takes $K$ parallel CUSUM algorithms as CD$(\cdot,\cdot)$ in CD-UCB. Since our CUSUM algorithm needs to estimate the mean first, we let the policy finish the estimation with $M$ observations as soon as possible. That is why we introduce the countdown timers $cnt(\cdot)$ in Algorithm~\ref{alg:CUSUM-UCB}. The performance of CUSUM-UCB depends on the parameters $\alpha$ and $h$. We discuss the joint choices of $\alpha$ and $h$ in Sections~\ref{sec:analysis} and~\ref{sec:simul}.
\begin{algorithm}[tb]
\caption{CUSUM-UCB}
\label{alg:CUSUM-UCB}
\begin{algorithmic}
\REQUIRE time horizon $T$, parameters $\alpha$, $\epsilon$, $M$ and $h$
\STATE Initialize $\tau_i=1$ and $cnt(i)=M$ for each $i\in\mathcal{K}$.
\FOR{$t$ {\bfseries from} $1$ {\bfseries to} $T$}
\STATE{Update according to equations (\ref{eqn:update}).}
\IF{$cnt(i)>0$ for some $i\in\mathcal{K}$}
\STATE $I_t=i$; $cnt(i)=cnt(i)-1.$
\ELSE
\STATE{Update $I_t$ according to equation (\ref{eqn:It}).}
\ENDIF
\STATE{{\blue Play arm $I_t$ and observe $X_t(I_t)$.}}
\IF{CUSUM$(I_t,X_t(I_t))==1$}
\STATE{$\tau_{I_t}=t+1$; $cnt(I_t)=M$; reset CUSUM$(I_t,\cdot)$.}
\ENDIF
\ENDFOR
\end{algorithmic}
\end{algorithm}

\section{Simulation Parameters}\label{app:simul}
\begin{table}[H]
  \centering
  \caption{
  Parameter setting in the simulation
  }
  \label{table:simul}
  \tabcolsep 3pt
\begin{center}
    \begin{tabular}{|c|c|c|c|c|c|c|c|}    
    \hline
   Environment & $K$ & $T$ & $\gamma_T$ & $\epsilon$
    & $M$  & $h$ & $\alpha$\\
    \hline
    Flipping &$2$ & $10^5$ & $2$ & $0.1$
    & $100$ & $50$ & $0.001$\\
    \hline
    Switching &$5$ & $10^6$ & $10$ & $0.1$
    & $100$ & $20$ & $0.01$\\
    \hline
    Yahoo! 1 & $5$& $5\times10^5$ & $32^{\dag}$ & $0.005$
    & $100$ & $200$ & $0.024$\\
    \hline
    Yahoo! 2& $100$ & $2\times10^6$ & $216^{\dag}$ & $0.005$
    & $100$ & $200$ & $0.024$\\
    \hline

  \end{tabular}
\end{center}
  \begin{tablenotes}
	\item[a] $\dag$: {\blue We count breakpoints when the difference in
          mean rewards is greater than $\epsilon=0.005$.}
  \end{tablenotes}
\end{table}

\end{document}